\documentclass[dvipsnames]{article}

\usepackage{arxiv}

\usepackage{amsmath,amsfonts,bm}

\def\eqref#1{equation~\ref{#1}}

\def\1{\bm{1}}

\DeclareMathAlphabet{\mathsfit}{\encodingdefault}{\sfdefault}{m}{sl}
\SetMathAlphabet{\mathsfit}{bold}{\encodingdefault}{\sfdefault}{bx}{n}

\newcommand{\R}{\mathbb{R}}

\newcommand{\sigmoid}{\sigma}
\newcommand{\softplus}{\zeta}

\newcommand{\softmin}{\mathrm{softmin}}
\newcommand{\logit}{\mathrm{logit}}

\DeclareMathOperator*{\argmax}{arg\,max}
\DeclareMathOperator*{\argmin}{arg\,min}

\usepackage[utf8]{inputenc} 
\usepackage[T1]{fontenc}    
\usepackage{hyperref}       
\usepackage{url}            
\usepackage{makecell}
\usepackage{booktabs}       
\usepackage{multirow}       
\usepackage{amsfonts}       
\usepackage{nicefrac}       
\usepackage{microtype}      
\usepackage{lipsum}
\usepackage{graphicx}
\graphicspath{ {./images/} }

\usepackage{amsmath}
\usepackage{amssymb}
\usepackage{mathtools}
\usepackage{amsthm}

\usepackage{microtype}      

\usepackage{tikz}
\usetikzlibrary{matrix}
\usetikzlibrary{calc}

\usetikzlibrary{fit, arrows.meta, positioning}
\usetikzlibrary{decorations.pathreplacing} 
\usepackage{float}

\usepackage{amssymb}
\usepackage{mathrsfs}
\usepackage{amsthm}
\newtheorem{proposition}{Proposition}
\theoremstyle{remark}
\newtheorem{remark}{Remark}

\usepackage{mathtools}
\usepackage{verbatim}
\usepackage{tabularx}
\usepackage{tcolorbox}

\newtheorem{definition}{Definition}

\definecolor{cobalt}{rgb}{0.02, 0.29, 0.65}
\definecolor{redUnipd}{RGB}{155, 0, 20}
\definecolor{forestgreen}{RGB}{10,210,34}

\title{Context-Selective State Space Models:\\ Feedback is All You Need}

\author{
  Riccardo Zattra \quad
  Giacomo Baggio \quad
  Umberto Casti \quad
  Augusto Ferrante \quad
  Francesco Ticozzi \\
  Department of Information Engineering \\
  University of Padova \\
  Padova, 35131, Italy \\
  \texttt{\{riccardo.zattra, baggio, castiumber, augusto, ticozzi\}@dei.unipd.it}
}

\begin{document}

\maketitle

\begin{abstract}
Transformers, powered by the attention mechanism, are the backbone of most foundation models, yet they suffer from quadratic complexity and difficulties in dealing with long-range dependencies in the input sequence. Recent work has shown that \textit{state space models} (SSMs) provide a promising alternative. In this paper, we introduce the {\bf COFFEE} (COntext  From FEEdback) model, a novel time-varying SSM that incorporates \textit{state feedback} to enable context-dependent selectivity, {while still allowing for parallel implementation.} 
This idea allows the model to regulate its dynamics based on the context
described by the internal state, which embodies a compact representation of the input history. State feedback allows COFFEE to improve its ability to capture long-range dependencies:
on the induction head task, it achieves near-perfect accuracy with two orders of magnitude fewer parameters and training sequences compared to S6 (the SSM of Mamba). On MNIST, COFFEE largely outperforms S6 within the same architecture, reaching  97\% accuracy with only 3585 parameters. These results showcase the role of state feedback as a key mechanism for building scalable and efficient sequence models.
\end{abstract}

\section{Introduction}
The {\em transformer} based on the celebrated {\em attention mechanism}, \cite{Transformer}, is the key of a recent groundbreaking improvement in generative Artificial Intelligence (AI), \cite{BERT,GPT4,Deepseek-r1,Gemini,EdgeFormer}. 
Yet, the attention mechanism suffers from some limitations due to a quadratic time complexity with respect to the length of the input sequence and severe difficulties in capturing long-range dependencies.
Several variations have been proposed to mitigate this problems, \cite{longformer,BigBird,Reformer,Performers,axialattention}. 
A completely different and extremely promising approach is the use of linear State-Space Models (SSMs): introduced in the 1960's, they quickly revolutionized control and automation engineering \cite{kailath-book} and many neighboring fields. In fact, a significant stream of valuable contributions have recently appeared that replace the attention mechanism with SSMs \cite{S5_model,diagonal_SSM,Strucutured_SSM,MAMBA,LRU,hasani2023liquid,dao2024transformers,jiang2024slot,muca2024theoretical,parnichkun2024statefree,rusch2025oscillatory,nishikawa2025state}. 
In \cite{zancato2024BMOJO}  a unifying framework is presented which embeds the transformer and many SSMs into the general setting of stochastic realization, 
\cite{LP-book};  In this framework a more general model is also presented which couples the advantages of the transformer short-term eidetic memory with those
of the long-term fading memory of SSMs. 
The connections between LLMs and control systems has been thoroughly investigated. In particular, controllability is analyzed in \cite{soatto2023tamingaibotscontrollability},
observability in 
\cite{liu2024meaningsfeelingslargelanguage} and stability in \cite{marchi2024heatdeathgenerativemodels}.

The starting point of this paper is the so called  \textbf{S6} SSM which lies at the core of the so-called  \textbf{Mamba} architecture \cite{MAMBA}.\footnote{The evolutions Mamba-2 \cite{MAMBA-2}, and Mamba-3 are powered by similar SSMs.}
The latter exhibits great capabilities and is able to outperform the state-of-the-art transformer in challenging tasks, including the Long Range Arena (LRA),  see \cite{MAMBA} and \cite{SSMControlOverview}. The reason for this success lies in the fact that 
the {\em state} is, by definition, the mathematical object that encodes all the relevant information on the past evolution of a system needed to predict its future behavior\footnote{See Section \ref{sec:background_on_ssm} and Appendix \ref{appendix:useful_concepts_on_ssm} for more details on this.}.  Of course, the main challenge is to derive a universal class of state-space models whose parameters can be learned for each specific task. The S6 module captures, or rather approximates, this universality by resorting to (linear) time-varying models. 

In this paper, we propose 
a novel class of SSMs that we call COFFEE (COntext From FEEdback) which is endowed with feedback capabilities. Dynamical systems endowed with feedback loops are at the basis for the large majority of complex natural and artificial phenomena, as they enable a system to adapt its behavior based on its current state, which, as previously recalled, contains all the relevant information about the past of the sequence under analysis. In this way, we are able to induce \textbf{context-selectivity} in the behavior of COFFEE: a property of paramount importance, which appears to yield substantial performance improvements in model performance.
With respect to S6, COFFEE  has another advantage (which could also be implemented in the S6 model): by exploiting a change of basis in the state-space representation of the model, we are able to reduce appreciably the number of learnable parameters, offering an obvious edge in terms of both numerical implementation and mechanistic interpretability.

We tested COFFEE in several versions of the induction head task and in the MNIST benchmark. Across these tasks, our model consistently outperforms S6 along multiple dimensions. In particular, COFFEE not only achieves substantially higher output accuracy, but also requires fewer learnable parameters and significantly smaller training datasets to reach convergence. 
The closed-loop COFFEE model is obtained by a nonlinear state-feedback action on a time-varying linear model. In general, this kind of structure leads to computational drawbacks because it is not amenable to parallelization and hence cannot take advantage of the highly parallel architecture of modern GPUs.
In COFFEE, however, the feedback structure is constrained in such a way that,
on the one hand, the performance degradation (with respect to the case of a more general feedback structure) is negligible and, on the other, the Jacobian of the state transition function is guaranteed to be diagonal. As a consequence, we can leverage recent results in \cite{gonzalez2024towards} to obtain scalable and highly efficient parallel training.

Finally, we mention that \cite{farsang2025scaling} proposes a biologically-inspired nonlinear SSM (LrcSSM) with diagonal Jacobian. While LrcSSM relies on intrinsically nonlinear state- and input-dependent dynamics, COFFEE restricts nonlinearity to a diagonal state-feedback gate that modulates a fixed linear SSM backbone, resulting in a more constrained and parameter-efficient model.

The remainder of this paper is organized as follows.
In Sec.~\ref{sec:background_on_ssm} we give a brief background on SSMs, with further details provided in Appendix \ref{appendix:useful_concepts_on_ssm}. In Sec.~\ref{sec:the_proposed_model}, we present our model along with its mathematical motivations. In Sec.~\ref{sec:experiments}, we assess the effectiveness of COFFEE by comparing it with S6 \cite{MAMBA} on the induction head and MNIST tasks. In Sec.~\ref{sec:theoretical_insights}, we provide some theoretical insights, based on an analysis made on a simplified version of the induction head task, whose details are presented in Appendix \ref{appendix:IH_task}. We summarize our findings in Sec.~\ref{sec:conclusion}, where we also outline the next steps in the development and testing of the model.

\section{SSMs and Sequence Modeling}

\label{sec:background_on_ssm}

\subsection{Linear, discrete SSM}
State space models are powerful tools to model, analyze and control dynamical systems
\cite{kailath-book}. These are models in which the relation between the input or control sequence $u(k) \in \mathbb{R}^{m}$  and the sequence $y(k) \in \mathbb{R}^{p}$ associated to the quantities of interest for the system at hand and called the {\em output}, is mediated by a 
signal $x(k) \in \mathbb{R}^n$, which is called the \emph{state} of the system. For more details, see Appendix \ref{appendix:useful_concepts_on_ssm}. 
In this paper, we leverage the class of linear, time-varying, discrete-time 
systems with finite-dimensional state space.
The models of this class are described by a pair of vector equations of the form 
\begin{equation}
\label{for:state_space_model_chap_5}
\begin{aligned}
    x(k) &= A(k) x(k-1) + B(k) u(k) \\
    y(k) &= C(k) x(k)
\end{aligned}
\end{equation}
where the matrix $A(k) \in \mathbb{R}^{n \times n}$ is called \emph{state transition matrix}, $B(k) \in \mathbb{R}^{n \times m}$ is called \emph{input matrix}, $C(k) \in \mathbb{R}^{p \times n}$ is called \emph{output matrix}. From (\ref{for:state_space_model_chap_5}) it follows that, given the state $x(k)$ at the present ``time'' $k$, the past of the input is irrelevant for the future of the output sequence. This key feature of the state is known as the {\em separation property}, and is the fundamental reason why the state functions as memory. The state can then be considered as a natural representative of the {\em context} set by the past data, and will be exploited in our setting in selecting the dynamics conveniently.

\subsection{Relevance of SSMs to Sequence Modeling}
In the context of sequence modeling, a state-space model as in (\ref{for:state_space_model_chap_5}) is used as a black-box in foundation models. The exogenous input $u(k)$ represents the $k$-th input token fed into the model, the state $x(k)$ stores relevant information about the tokens up to the $k$-th one, while $y(k)$ is the $k$-th output: for generative or prediction models $y(k)$ is used to predict $u(k+1)$; more generally, $y(k)$ is the signal used to solve the task at hand.
This~offers~several~advantages:
\\
\textbf{Efficiency}: SSMs are based on linear recurrences that enable efficient inference and allow training to be parallelized via FFT-based or associative scan algorithms.
\\
\textbf{Long-range dependencies}: By construction, the state of the system contains a compressed representation of the past history of the system: the state is capable of ``remembering'' all previously seen tokens in a compact form;
\\
\textbf{Interpretability~and~structure}: The matrices $A(k)$, $B(k)$, $C(k)$ explicitly define the system dynamics, providing both structure and, ideally, interpretability.

Recent neural architectures such as S4 \cite{Strucutured_SSM} and S6 \cite{MAMBA} leverage these ideas, combining them with modern training techniques and nonlinearities to build powerful sequence models. In particular, S4 uses a time-invariant model, namely a model of the form (\ref{for:state_space_model_chap_5}) where, however, the matrices $A,B,C$ are independent of $k$,  while the main novelty of S6 is that the matrices of the SSM are input-dependent and hence time-varying.

In general, the matrices $A,B,C$ are structured (typically $A$ is diagonal) and are learned during training.

\section{COFFEE: Towards Context-Selective SSMs}

\label{sec:the_proposed_model}

In this section, we introduce a novel state space model which we call {COFFEE} (COntext From FEEdback). S6 \cite{MAMBA} enables selectivity conditioned on the current token by adapting the system dynamics to the current input. COFFEE replaces this \emph{token-based selectivity} with \emph{context-based selectivity}, achieved through a feedback mechanism that modulates the dynamics based on the {\em state} of the model.

Both COFFEE and S6 are implemented within the core module architecture described in Appendix~\ref{appendix:core_module_architecture}, which maps sequences over a finite vocabulary $V$ into embeddings in $\mathbb{R}^D$, and processes each embedding feature with a dedicated single-input single-output (SISO) state-space model. To motivate our modifications, we now briefly review the structure of the S6 SISO component (a more detailed analysis is available in Appendix~\ref{appendix:the_S6_model}).

\subsection{Background: The S6 Model}
\label{subsec:S6_overview}

As mentioned above, in S6 each feature $u(k)_i$ of the embedding $u(k)\in \mathbb{R}^D$ is handled independently by a dedicated SISO state-space model $\Sigma^{(i)}$. This design allows the architecture to scale across the embedding dimension by running $D$ parallel SISO models.  
The SSM for each feature is:
\begin{equation}
\label{eq:S6_model}
\begin{cases}
  x(k)^{(i)} = e^{A^{(i)}\Delta_i(k)} x(k-1)^{(i)} +\ \big(A^{(i)}\big)^{-1}\!\big(e^{A^{(i)}\Delta_i(k)}-I \big) B(k)\, u(k)_{i} \\
  y(k)^{(i)} = C(k)\, x(k)^{(i)}
\end{cases}
\end{equation}
where $x(k)^{(i)} \in \mathbb{R}^n$ is the hidden state of the $i$-th feature, $A^{(i)}=\mathrm{diag}(\lambda_0^{(i)},\dots,\lambda_{n-1}^{(i)}) \in\mathbb{R}^{n\times n}$ is a diagonal matrix (a suitable learning implementation is adopted to avoid vanishing \smash{$\lambda_{j}^{(i)}$}), and $u(k)_i$ denotes the $i$-th feature of the embedding vector $u(k)$.  
The input-to-state and state-to-output mappings are $B(k) = W_B u(k)$ and $C(k) = (W_C u(k))^\top$, 
with $W_B, W_C \in \mathbb{R}^{n\times D}$ shared across all features. The step sizes $\Delta_i(k)$ are determined as the components of the vector
\begin{equation}\label{eq:token-based-gating}
\Delta(k) = \softplus(W_D u(k)),
\end{equation}
where $W_D \in \mathbb{R}^{D\times D}$ and $\softplus(\cdot)$ denotes the softplus function. 
Notice that, while $B(k)$ is the same for each $i$, the overall input matrix \smash{$(A^{(i)})^{-1}\! (e^{A^{(i)}\Delta_i(k)}-I ) B(k)$} is indeed feature-dependent. 
A distinctive aspect of S6 is its token-based selectivity: the parameters $B(k)$, $C(k)$, and $\Delta(k)$ depend directly on the current token $u(k)$. This enables the model to dynamically amplify or suppress contributions from different inputs, effectively filtering out irrelevant symbols while focusing on meaningful ones. Such input-conditioned adaptation is particularly advantageous in sequential domains (e.g., speech, where disfluencies carry little meaning). This token-dependent, time-varying structure is widely regarded as the core innovation that underpins the strong empirical performance of Mamba and its descendants.

\subsection{The COFFEE  Model}
\label{sec:FS6}

The central novelty of COFFEE is the use of \emph{state feedback} to achieve context-based selectivity. Unlike S6, where the update gates $\Delta_i(k)$ depend only on the current input token, COFFEE computes them as functions of the internal state, allowing the model to adapt its dynamics based on the accumulated context. It is well known that open-loop systems are fragile and sensitive to model uncertainties and disturbances, whereas closed-loop feedback ensures robustness; COFFEE exploits this principle to achieve context-based selectivity, leading to more stable and~consistent~generative~behavior.

Beyond feedback, we further refine the S6 formulation through two steps: (i) linearizing the exponential dynamics to make the role of $\Delta(k)$ more interpretable, and (ii) eliminating parameter redundancy via changes of basis in the input and state spaces. A complete derivation of COFFEE  is presented in Appendix~\ref{appendix:derivation_of_our_model}; below we summarize the key steps leading to the proposed model.

\paragraph{Step 1. Simplified Dynamics via Linearization.} 
In S6, the step size $\Delta_i(k)$ controls the tradeoff between retaining past state and incorporating new input. However, its action is embedded within matrix exponentials, which complicates direct analysis. To make this dependence more explicit and obtain a more interpretable formulation, we approximate $e^{A^{(i)}\Delta_i(k)}$ with its first-order Taylor expansion around $\Delta_i(k)=0$. Furthermore, we replace the matrices $B(k)$ and $C(k)$ with feature-dependent, time-invariant ones. 
The choice of feature-dependent matrices is motivated by the fact that, due to discretization, the input-to-state mapping in (\ref{eq:S6_model}) is already feature-dependent, while we additionally assign feature-dependent $C^{(i)}$ to maintain symmetry. Instead, time-invariance allows selectivity be governed solely by the step size $\Delta_i(k)$. 
This yields the dynamics:
\begin{equation}
\label{eq:linearized_model}
\begin{cases}
  x(k)^{(i)} = \big(I + A^{(i)}\Delta_i(k)\big)\,x(k-1)^{(i)}  +\ \Delta_i(k) B^{(i)} u(k)_{i} \\
  y(k)^{(i)} = C^{(i)} x(k)^{(i)}
\end{cases}
\end{equation}
where $B^{(i)}\in \mathbb{R}^{n\times 1}$ and $C^{(i)}\in \mathbb{R}^{1\times n}$ are time-invariant, feature-dependent matrices. 
By replacing exponentials with affine dependence on $\Delta_i(k)$, the simplified model in (\ref{eq:linearized_model}) preserves the essential gating mechanism of S6 while rendering the dynamics simpler and more amenable to interpretation.

\paragraph{Step 2. Feedback for Context-Based Selectivity.}  
A key strength of S6 lies in its token-based selectivity: the update gates $\Delta_i(k)$ are computed as functions of the current input token $u(k)$, so that the model can suppress irrelevant inputs and amplify informative ones. However, token-based selectivity is inherently limited as identical tokens may play different roles depending on context.
As a simple example, consider approximating the number ``111.11'', where each digit ``1'' is represented by the same token. In this case, the model must treat different occurrences of ``1'' differently depending on their position.
To address this limitation, we observe that the hidden state $x(k-1)^{(i)}$ of each subsystem $\Sigma^{(i)}$ already encodes a compressed representation of the full history up to step $k-1$, and thus provides a natural source of contextual information. 
We therefore replace the input-dependent gating mechanism with a state-dependent one. In addition, we extend the scalar gate $\Delta_i(k)$ in (\ref{eq:linearized_model}) to a vector-valued gate to enable efficient parallel training, as discussed in Sec.~\ref{sec:parallelization}. Concretely,
we replace the scalar token-based gating (\ref{eq:token-based-gating}) with a \emph{vector-valued, state-dependent gating rule}:
\begin{equation}\label{eq:Delta-feedback}
\Delta_i(k) = \sigmoid\big(w_D^{(i)} \odot x(k-1)^{(i)}\big),
\end{equation}
where $w_D^{(i)} \in \mathbb{R}^n$ is a learnable parameter vector, $\odot$ denotes the Hadamard product, and $\sigmoid(\cdot)$ is the sigmoid function applied componentwise to vectors.\footnote{We use the sigmoid rather than the softplus, since its bounded range keeps $\Delta_i(k)$ limited, helping maintain the validity of the Taylor approximation in (\ref{eq:linearized_model}).} By doing so, each subsystem adapts its update gate from its past history rather than the raw input token, enabling context-based selectivity: identical tokens can be filtered differently depending on their role in the sequence. Concretely, the gating vector $\Delta_i(k) \in \mathbb{R}^n$ enters the dynamics in (\ref{eq:linearized_model}) as a diagonal state feedback matrix, which modulates the contribution of the past state through
$
(I + A^{(i)}\mathrm{diag}(\Delta_i(k)))x(k-1)^{(i)},
$
and the input through
$
\mathrm{diag}(\Delta_i(k)) B^{(i)}u_i(k).
$
The resulting subsystem $\Sigma^{(i)}$ is therefore context-adaptive, taking the form
\begin{equation}
\label{eq:model_with_feedback}
\begin{cases}
  x(k)^{(i)} =
    (I+A^{(i)}\mathrm{diag}(\Delta_i(k)))\,x(k-1)^{(i)}  +\, \mathrm{diag}(\Delta_i(k)) B^{(i)}\,u(k)_i, \\
  y(k)^{(i)} = C^{(i)} x(k)^{(i)}.
\end{cases}
\end{equation}

\begin{remark}[Parameter efficiency]   
Compared to S6, the proposed feedback formulation is significantly more parameter-efficient: whereas S6 requires a dense $W_D\in\mathbb{R}^{D\times D}$, our version replaces it with $D$ vectors $w_D^{(i)}\in\mathbb{R}^n$, i.e., $Dn$ parameters in total.  
Since typically $n\ll D$ (e.g., $n=16$, $D=2048$ in Mamba\footnote{The reported numbers are taken from the minimal PyTorch implementation of Mamba (\url{https://github.com/johnma2006/mamba-minimal}), numerically equivalent to the official release (\url{https://github.com/state-spaces/mamba/tree/main}).}), this corresponds to a reduction from $4.2$M to just $32$K gating parameters.
\end{remark}

\paragraph{Step 3. Eliminating Redundancy via Change of Basis.}  

As a final step, we note that the parameterization of $(B^{(i)}, C^{(i)})$ in (\ref{eq:model_with_feedback}) exhibits parameter redundancy: through suitable changes of basis (corresponding to simple rescalings) the same input-output behavior can be preserved while reducing the number of parameters. This fact is formalized in the following proposition, whose proof is deferred to the Appendix \ref{proof-prop-changebasis}, where a few comments are also provided on the proposition assumptions.

\begin{proposition}[Redundancy elimination by change of basis]\label{prop:redundancy_elimination}
Assume that at least one embedding vector has all the entries that are non-zero and that all the entries of the $B^{(i)}$ are non-zero. For the family of subsystems $\{\Sigma^{(i)}\}_{i=0}^{D-1}$ defined in (\ref{eq:model_with_feedback}),  there exist changes of basis in the input and state spaces such that the structure of (\ref{eq:model_with_feedback}) is preserved and:\\ 
(i) all input-to-state matrices $B^{(i)}$ can be fixed to the constant vector $\mathbf{1}=[1,\dots,1]^\top$, with their scaling absorbed into the corresponding $C^{(i)}$, and\\  
(ii) one embedding vector can be normalized to $\mathbf{1}$.  
\end{proposition}

The transformations in Proposition~\ref{prop:redundancy_elimination} preserve both the input-output map and the diagonal structure of $A^{(i)}$. Consequently, the input-to-state matrices $B^{(i)}$ are redundant, as their effect can always be absorbed into $C^{(i)}$, and one embedding vector can be fixed to $\mathbf{1}$. This reduces the parameter count by $nD+D$ without loss of expressivity, yielding a leaner architecture.

\begin{figure*}[t]
\centering
\includegraphics[scale=0.775]{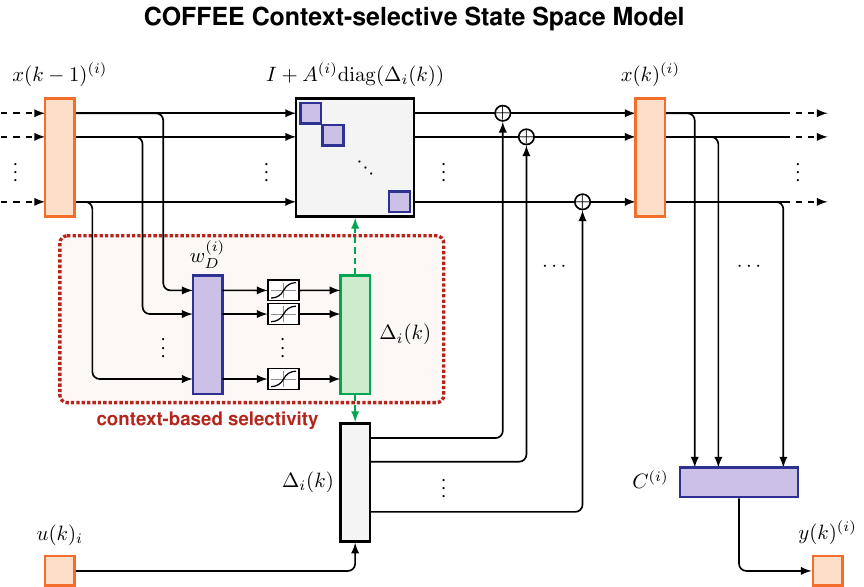}
\vspace{0.25cm}
\caption{\textbf{Block diagram of the $i$-th feature subsytem of COFFEE.} 
Orange denotes the model signals, blue the learnable parameters, 
and green the selectivity or gating parameter, which is obtained from the past state through (\ref{eq:Delta-feedback}).  
By feeding this parameter back into the dynamics, the model forms a closed-loop mechanism that provides context-based selectivity, enabling identical tokens to be processed differently depending on their role in the sequence.}
\label{fig:diagram}
\end{figure*}

\paragraph{Final COFFEE Model.}  
Combining Steps 1-3, the resulting COFFEE  subsystem for feature $i$ takes the form
\begin{equation}
\label{eq:final_model}
\begin{cases}
x(k)^{(i)} = (I+A^{(i)}\mathrm{diag}(\Delta_i(k)))\,x(k-1)^{(i)} + \Delta_i(k)\,u(k)_i,\\
  y(k)^{(i)} = C^{(i)} x(k)^{(i)} ,
\end{cases}
\end{equation}
where the state-dependent gates $\Delta_i(k)$ are given in (\ref{eq:Delta-feedback}). 
Here, the learnable parameters are the diagonal elements $\{\lambda_k^{(i)}\}_{k=0}^{n-1}$ of the matrix $A^{(i)}$, the output matrix $C^{(i)}\in\mathbb{R}^{1\times n}$, and the feedback vector $w_D^{(i)}\in\mathbb{R}^n$. To ensure numerical stability during training and inference, we enforce stability for the matrix $I + A^{(i)}$. A schematic representation of COFFEE is shown in Fig.~\ref{fig:diagram}.

\begin{remark}[Model with output filtering]\label{remark:output_filtering}
Notice that the model in (\ref{eq:final_model}), can be extended by introducing additional filtering mechanisms at the output level.
In particular, we can transform the output equation in:
\[
y(k)^{(i)} = (C^{(i)}\gamma(k)^{(i)}) x(k)^{(i)}
\]
where $\gamma(k)^{(i)} \in \mathbb{R}$ is given by:
\[
\gamma(k)^{(i)} = \sigma((w_{\gamma}^{(i)})^{\top}x(k)^{(i)})
\]
and $w_{\gamma}^{(i)}\in \mathbb{R}^{n}$ are learnable parameters. This extension introduces an additional $nD$ learnable parameters.

Notably, since the output equation is static, this generalization does not  compromise the possibility, discussed in the next section, of parallelizing the model.
This remains true even if other nonlinearities are added~to~the~output~equation.
\end{remark}

\section{Parallelization}\label{sec:parallelization}
Architectural parallelization is a fundamental requirement to fully exploit the computational capabilities of modern GPUs, including Tensor and CUDA cores. However, the model described in Equation (\ref{eq:final_model}) does not naturally admit an efficient parallel implementation. Indeed, computing the state $x(k)$ requires access to the previous state $x(k-1)$, thus pointing to an intrinsic sequential dependency. 

Moreover, when implemented using standard automatic differentiation frameworks (e.g., PyTorch, JAX), this approach leads to the construction of a large computational graph. Such a graph not only incurs significant memory overhead but may also cause vanishing gradients or numerical instabilities during gradient backpropagation.

To overcome these limitations, we adopt a parallelization strategy originally introduced by \cite{DEER_lim} and subsequently refined by \cite{gonzalez2024towards}. In \cite{DEER_lim}, an algorithm known as DEER is developed to address both the discrete-time and continuous-time formulations. When specialized to the discrete-time recurrent neural network setting, such framework allows inference to be reformulated as the solution of a fixed-point problem, which can then be efficiently solved using Newton’s method. 
To use this theory in our setting, we write the state equation of (\ref{eq:final_model}) in the form 
\[
x(k) = f_k(x(k-1))
\]
where $x(k)\in\mathbb{R}^{\tilde{n}}$ may denote each single 
component $x(k)^{(i)}$ (in which case $\tilde{n}=n$), or the whole state obtained by stacking the $x(k)^{(i)}$ on top of each others (in which case $\tilde{n}=Dn$);
in fact, depending on the hardware and on the values of $n$ and $D$, it may be convenient to consider each $\Sigma^{(i)}$ separately or to group them together as a unique SSM.
The function $f_k$ represents the nonlinear state transition function and its dependence on $k$ also accounts for the contribution of $u(k)$. Given a collection of candidate states $\{x(k)\}_{k=1}^{L}$, where $L$ denotes the sequence length, we define the~{\em residual}~as:
$$
r(x_{1:L}) \;:=\;  [x(1)-f_1(x(0)),x(2)-f_2(x(1)),...,  x(L)-f_L(x(L-1))]  
$$
where $x_{1:L}$ denotes the tentative overall trajectory of the system starting from a given initial state $x(0)$. The true trajectory originating from $x(0)$ is the unique collection of states $x^{*}_{1:L}$ satisfying
$r(x^{*}_{1:L}) = 0.$
To find $x^{*}_{1:L}$, Newton's method is employed. It can be shown that doing so is equivalent to computing the trajectory of a first-order linear recurrence, which can in turn be executed in parallel using the parallel associative scan \cite{blelloch1990prefix}. However, in order to apply Blelloch's algorithm, it is necessary to store $\frac{\partial f_{i}}{\partial x}(x(i-1))$ for $i = 1, \dots, L-1$. If the Jacobians are dense, the scalability is limited because the total computational work scales as $\mathcal{O}(L\tilde{n}^{3})$ and the memory requirement scales as $\mathcal{O}(L\tilde{n}^{2})$. For mathematical details, we refer the reader to Appendix \ref{Appendix:parallelization}.
For this reason, we resort to the refined strategy proposed in
\cite{gonzalez2024towards}, which improves both scalability and numerical stability. Regarding scalability, instead of storing the full Jacobians, only their diagonals are considered, resulting in the so-called quasi-DEER algorithm, with total computational work $\mathcal{O}(L\tilde{n})$ and memory scaling as $\mathcal{O}(L\tilde{n})$. Concerning numerical stability, the problem was reformulated as an optimization problem, recognizing that a Gauss-Newton method can be effectively used to minimize the objective. To further enhance stability, the Levenberg-Marquardt algorithm was adopted. This algorithm can be interpreted as a Gauss-Newton method equipped with a trust region to improve robustness, see \cite{gonzalez2024towards} for further details.

Building on the previous considerations, our model has been designed so that the Jacobians are inherently diagonal, making quasi-DEER an exact implementation rather than an approximation of DEER. Specifically, in our model, $A^{(i)}$ is diagonal, and each component of $\Delta_i(k)$ depends solely on the corresponding component of $x(k-1)^{(i)}$.  Consequently, the state dynamics in (\ref{eq:final_model}) decomposes into $\tilde{n}$ independent scalar recursions. This further implies that the state-transition Jacobian is diagonal at every time step.

\section{Experiments}

\label{sec:experiments}
We report numerical experiments on two primary tasks: Induction Head (including variants) and MNIST. Implementation details are provided in Appendix~\ref{Appendix:result_IH} and Appendix~\ref{appendix:core_module_architecture} for the Induction Head, and in Appendix~\ref{Appendix:MNIST_arch_and_results} for MNIST.

\subsection{Results on the Induction Head task}
In the following, we present results from a series of experiments designed to solve the Induction Head (IH) Task, originally introduced in \cite{InductionHead} and adopted as one of the motivating tasks in Mamba \cite{MAMBA}. This task was proposed to evaluate in-context learning capabilities of modern deep learning architectures. An explanation of the induction head task is reported in Appendix \ref{appendix:IH_task}. In what follows, the adopted vocabulary is $V=\{1,2,3,4,5,6,7\}$.

We first compare the performance of COFFEE and S6 on the IH task with $L_{\text{seq}}=16$, $L_{\text{tri}}=1$ and $L_{\text{tar}}=1$, where $L_{\text{seq}}$, $L_{\text{tri}}$ and $L_{\text{tar}}$ are the lengths of the whole sequence, the trigger and the target,  respectively.
As shown by the results in Table \ref{table1}, COFFEE is able to reach $0.99+$ accuracy in just one epoch, namely with only 5'120'000 training sequences. The S6 performance is obtained by running the training algorithm for 100 epochs (512'000'000 training sequences).
\begin{table}[H] 
  \centering{
  \caption{Comparison between S6 and COFFEE on the IH task. Different learning rate
  $\eta$ were adopted for both S6 and COFFEE, the best ones are reported.
  For both models the state dimension is $n=8$ and the embedding dimension is $D=16$.}
  \begin{tabular}{lrrrrrr}
    \toprule
    Model & \makecell[c]{$\eta$}& \# par. & Epochs & Loss &  Acc. \\
    \midrule
    S6   &  0.003 & 784 & 100& 0.852 &  0.68  \\
    COFFEE & 0.01 & 512 & 1& 0.000 & 0.99+  \\
    \bottomrule
  \end{tabular}
  \label{table1}}
\end{table}

To test the limits of COFFEE we add an extra difficulty by reducing the state and embedding dimensions. Given the performance of S6 in the previous simplified scenario, we do not provide a comparison with COFFEE in more complicated versions of the IH task reported in Table \ref{table1}.
\begin{table}[H]
  \centering
  \caption{COFFEE performance on the IH task with reduced state and embedding dimensions.
  The learning rate is $\eta=0.01$. The  embedding dimension is
  denoted by $D$, the state dimension by $n$ and \# par.~denotes the number of learnable parameters.
  }
  \begin{tabular}{lcccccr}
    \toprule
    Model &  $n$ &$D$  & \# par. & Loss &  Accuracy \\
    \midrule
    COFFEE &  1&8  & 96  & 0.004 & 0.99+  \\
    COFFEE &  1&4  & 48  & 0.002 & 0.99+  \\
    COFFEE &  1&2  & 24  & 0.035 & 0.99  \\
    COFFEE &  1&1  & 12  & 1.793 & 0.16  \\
    \bottomrule
  \end{tabular}
  \label{table2}
\end{table}

We see that COFFEE achieves near perfect accuracy with only 24 parameters which is a remarkable performance.

To further assess the capabilities of COFFEE, we test it on a variation of the IH task. The standard implementation of the IH task, as explained in Appendix \ref{appendix:IH_task}, assumes a sequence with the following structure:\\[1mm]
\hspace*{15mm}$\text{noise}\ \| \ \text{trigger}\ \| \ \text{target}\ \| \ \text{noise}\ \| \ \text{trigger}$\\ [1mm]    
where $\|\ $ denotes sequence concatenation. We modify the above structure in:\\[1mm]
\hspace*{12mm}$ 
    \text{noise}\ \| \ \text{trigger}\  \| \ \text{noise}\ \| \ \text{target}\ \| \ \text{noise}\ \| \ \text{trigger}
$\\ [1mm] 
We evaluate our model in this scenario with $L_{\text{seq}}=16$, $L_{\text{tri}}=1$ and $L_{\text{tar}}=1$ and we obtain the results in Table \ref{table3}.
\begin{table}[H]
  \centering
  \caption{COFFEE model performance on a variation of the IH task.
  Here the state dimension is $n=8$, the embedding dimension is $D=16$ and the learning rate is $\eta=0.003$. The length of the noise subsequence between the trigger and target subsequences is denoted by $L_{\text{noise}}$.}
  \begin{tabular}{lccccr}
    \toprule
    Model &  $L_{\text{noise}}$ & Loss &  Accuracy \\
    \midrule
    COFFEE &  1 & 0.000 & 0.99+  \\
    COFFEE &  2 & 0.002 & 0.99+  \\
    \bottomrule
  \end{tabular}
  \label{table3}
\end{table}

We also evaluate COFFEE on the IH task under different trigger and target lengths: $L_{\text{tri}}=1$,  $L_{\text{tar}}=2$ and $L_{\text{tri}}=2$,  $L_{\text{tar}}=1$ while keeping $L_{\text{seq}}=16$. Results are in Table~\ref{table4}.
\begin{table}[H]
  \centering
  \caption{COFFEE model performance with different values of $L_{\text{tri}}$ and $L_{\text{tar}}$.
  Here the state dimension is $n=8$, the embedding dimension is $D=16$ and the learning rate is denoted by $\eta$.}
  \begin{tabular}{lccccc}
    \toprule
    Model &$\eta$&$L_{\text{tri}}$ & $L_{\text{tar}}$ & Loss &  Accuracy \\
    \midrule
    COFFEE & 0.01 & 1 & 2 &0.004 & 0.99  \\
    COFFEE & 0.003 & 2 & 1 &0.660 & 0.78  \\
    \bottomrule
  \end{tabular}
  \label{table4}
\end{table}

To conclude, we test our model on the IH task with increasing sequence lengths $L_{\text{seq}}=32,64,128,256$, $L_{\text{tri}}=1$ and $L_{\text{tar}}=1$. The results and settings are reported in Table \ref{table6}.
\begin{table}[H]
  \centering
  \caption{COFFEE model performance with increased sequence length.
  Here the state dimension is $n=8$, the embedding dimension is $D=16$ and the learning rate is   $\eta=0.01$. The length of the sequence is denoted by $L_{\text{seq}}$.}
  \begin{tabular}{lcccccc}
    \toprule
    Model
    &   $L_{\text{seq}}$
    & Loss &  Accuracy \\
    \midrule
    COFFEE & 32   & 0.000 & 0.99+  \\
    COFFEE & 64 & 0.001 & 0.99+  \\
    COFFEE & 128  & 0.000 & 0.99+  \\
    COFFEE & 256 & 0.001 & 0.99+  \\
    \bottomrule
  \end{tabular}
  \label{table6}
\end{table}

Overall, the experimental results show that COFFEE outperforms S6 across multiple IH settings, attaining higher accuracy while requiring significantly fewer parameters and training sequences. Given that S6 represents a strong baseline among selective SSMs, these findings highlight the potential of state-feedback mechanisms for improving performance on in-context learning tasks.

\textbf{Ablation studies.} COFFEE was derived through three main steps. First, we linearized S6; second, we applied a state-feedback; third, we performed a change of basis to eliminate redundancy. A natural question is then which of these three steps contributes the most to the overall performance. To address this question, Table \ref{table-ablation} reports experimental results on the IH task with $L_{\text{seq}}=16$, $L_{\text{tri}}=1$, and $L_{\text{tar}}=1$ considering  different ``partial" models.

\begin{table}[H]
  \centering
  \caption{Performance of the S6 model, COFFEE with linearization only as in Step 1, COFFEE obtained by applying Steps~1 and~3 but no state feedback ({\em COFFEE no fb.}), and the full model ({\em COFFEE final.}). To assess the importance of the nonlinearity in the state feedback (the sigmoid in~(\ref{eq:Delta-feedback})), we also tested the model without it; the corresponding results are reported as~\emph{COFFEE~lin.~fb.}. 
  }
  \begin{tabular}{lcccccr}
    \toprule
    Model &  $n$ &$D$  & \# par. & Loss &  Accuracy \\
    \midrule
    S6 &  8 & 16  & 784  & 0.852 & 0.69  \\
    COFFEE as in (\ref{eq:linearized_model})&  8& 16 & 784  & 0.906 & 0.68  \\
    COFFEE no fb. &  8&16& 512  & 0.788 & 0.74  \\
    COFFEE final (\ref{eq:final_model})&  8&16  & 512  & 0.000 & 0.99+  \\
    COFFEE lin. fb. &  8&16  & 512  & 1.845 & 0.16  \\
    \bottomrule
  \end{tabular}
  \label{table-ablation}
\end{table}

\begin{remark}
All experiments reported in this section were repeated multiple times, with~negligible~standard~deviation.
\end{remark}

\subsection{Results on the MNIST task}\label{sec:MNIST}
To further assess our model ability to capture dependencies between tokens, we tested our SSM on the MNIST dataset. For each model, we employ a {\em single state-space layer without feedforward}, with the same number of features. The adopted architecture and the hyperparameters setting are fully explained in Appendix \ref{Appendix:MNIST_arch_and_results}.  In Table \ref{table7} we report a summary of the results obtained: in the table, ``COFFEE + OF'' refers to the usual COFFEE model improved with filtering capabilities on the output (see Remark \ref{remark:output_filtering}). 
\begin{table}[H]
  \centering
  \caption{Performance comparison between COFFEE and S6 using the MNIST dataset.
  Here, we consider $100$ Epochs. We denote by $n$ the state dimension and by
  \# par. the overall number of learnable parameters of the model. Experiments were repeated ten times. }
  \label{tab:MNIST}
  \begin{tabular}{lcccccc}
    \toprule
    Model
    
    & $n$
    & \# par.
    & Loss &  Acc. & Std \\
    \midrule
    S6  & 2 & 5585 & 1.891 & 28.2\% &  0.1\\
    S6  & 16 & 10085 & 1.876 & 28.6\% &  0.1\\
    COFFEE  & 2 & 3385 & 0.107 & 96.6\% &  0.1 \\
    COFFEE + OF  & 2 & 3585 & 0.100 & 97.0\% & 0.1 \\
    \bottomrule
  \end{tabular}
  \label{table7}
\end{table} 
This experiment demonstrates the practical applicability of COFFEE in a scenario that is very different from those considered in the previous section, and show how the S6 performance is affected once the feedforward layer is removed. As discussed in \cite{MAMBA}, selectivity is a fundamental property of successful architectures for sequential modeling. In the previous section we established the relevance and efficiency of the selection mechanism introduced in COFFEE for  tasks in the induction head family; with the results summarized in Table~\ref{tab:MNIST}, we show how this embedded mechanism transfers to a completely different and more complex task, without incurring a prohibitive increase in model size.

In summary, we demonstrate that COFFEE is flexible and outperforms  S6  in very different tasks: for these reasons, we believe it deserves consideration as a building block in complex architectures, such as Mamba \cite{MAMBA}.

\section{Functioning principles of COFFEE}

\label{sec:theoretical_insights} 

The COFFEE model has been designed with a few basic, interpretable functioning principles in mind:
\\
{\bf P1:} The state represents the memory of the system, which is propagated to the next iteration by the state matrix. \\
{\bf P2:} The update gates $\Delta_i(k)$ select if and how much of the input $u(k)_i$ is memorized in the state dynamics.\\
{\bf P3:} Different areas of the state space allow the next encoded symbol to affect its future evolution differently (context selectivity).

In order to investigate whether the proposed model is indeed operating by exploiting these principles, we reduced the ``complexity''  of the induction head task as much as possible, so that we could geometrically represent the evolution of the state. In fact, even the simple architecture that solved the {inductive head} task with 24 parameters (see Section \ref{sec:experiments}), turned out to be impractical to study due to the lack of convenient visualization of the parameter dynamics.
The simplified inductive head task (IH0, described in detail in Appendix \ref{sec:reduction_IH}) considers sequences of four symbols chosen in $V=\{1,2,3\},$ where $1$ represents the trigger. The trigger appears two times in each sequence, one of which as the last symbol. To solve this problem, we chose embeddings in $\R^2,$ and a COFFEE model composed of two one-dimensional SSM, so that the evolution can be depicted in $\R^2$ as well. 

In order to highlight the role of the functioning principles described above, we introduce further model simplifications:
 For {\bf P1}, we choose   all $A^{(i)}=0$ in Equation (\ref{eq:final_model})), thus implementing a perfect memory. Furthermore, we also assume $C^{(i)}=1$ for $i=1,2$, so that the state values correspond to the outputs, and simplify interpretation. 
For {\bf P2}, we impose $w_D^{(i)}=1$, so that the state value remains the only relevant parameter. High positive values of the state components correspond to update gates close to one, and negative values bring the update gates close to zero.  
 For {\bf P3}, a state in the upper half-plane will cause the next input to ``save'' its second component (feature) in memory, and the right half-plane will memorize the first one. The first quadrant will save both, and the third quadrant neither.
With these restrictions in place, the only parameters left to learn are the embeddings.

We can now intuitively guess the general form of a successful embedding:
 In particular, the {\em trigger symbol} embedding shall move the state in a region of activation of the sigmoid that is high and ``neutral'' with respect to the embedding of the target symbols. Natural candidates are vectors on the bisector of the first quadrant.
The {\em embeddings of the  target symbols} shall
(i) be well distanced, in order to ensure distinguishability of the output; 
(ii)  move the state from the triggered position to an area that is closer to their position than the others, so that their symbol is chosen as the final output;
{(iii)} lower the activation of the sigmoid, and hence the update gates, so that further inputs will not cause the state to move too much towards the embedding of different symbols. Natural candidates are then large vectors in the third quadrant. 

We next verify that the {\em learned embeddings} for the IH task respect these principles. 
We trained the simplified model starting from a tentative embedding that satisfies the design principles sketched above, where:
$$
\vspace{-2mm}1  \rightarrow  [6,6]^{\top},\quad 
2  \rightarrow  [-10,-1]^{\top},\quad
3  \rightarrow  [-1,-10]^{\top}.
$$
From this initial condition, which does not always yield the correct results, the learning phase converges  quickly, and the learned model solves the task with $100\%$ accuracy. The learned solution corresponds to the embedding:
$$
1 \! \rightarrow  [5.4,5.3]^{\top},\, 
2 \! \rightarrow  [-10.4,-1.6]^{\top},
3 \! \rightarrow  [-1.5,-10.3]^{\top}
$$
The full trajectories of the state variables are depicted in Figure \ref{fig:trajectory_of_hand_crafted_sol} in the Appendix.  Their behavior shows the design principles {\bf P1-3} in action. It is worth noting that in this simple example, despite the existence of such optimal solution, the training phase does not always end successfully due to the extremely limited number of parameters to be optimized (6, corresponding to the embedding), and considering a suitable initialization is key\footnote{These issues do not emerge for the fully-parametrized model.}. Increasing the number of parameters quickly improves the training success, providing clear evidence of the beneficial effects of introducing redundancies in {number of trainable parameters}, despite being inefficient in terms of resources.
In this sense, {\bf P1-P3} represent both a way to interpret the behavior of the model, and a useful starting point to propose effective initializations, even in the presence of fewer redundant parameters.


\section{Conclusion}
\label{sec:conclusion}

In this work, we propose a new context-selective state-space model (COFFEE) in which a nonlinear state-feedback mechanism is designed to enhance model expressivity while remaining parameter-efficient and parallelizable. 
In the considered benchmarks (induction head and MNIST), COFFEE consistently outperforms S6  when the two models are instantiated within the same architectural setting. Ablation studies clearly identify the introduction of nonlinear feedback as the key source of such an advantage.

At this stage, COFFEE is still a proof of principle, since its performance in more complex benchmarks, such as the Long Range Arena, has not yet been tested. Extending COFFEE to these settings will naturally require a Mamba-like architecture composed of multiple COFFEE modules. Nonetheless, the strong accuracy achieved by a single COFFEE module with few learnable parameters provides encouraging evidence for the scalability of the proposed approach.

Building on these results, ongoing work is focused on integrating COFFEE into deeper architectures and conducting systematic comparisons with Mamba and other non–state-space models on more challenging and diverse benchmarks. Crucially, the numerical feasibility and computational efficiency of these extensions are supported by the structural properties of COFFEE: the resulting diagonal Jacobian ensures that training remains amenable to parallelization.

\newpage





\bibliography{references}
\bibliographystyle{unsrt}

\newpage
\appendix
\onecolumn
\section{Useful Concepts of State Space Models}  \label{appendixA}
\label{appendix:useful_concepts_on_ssm}
In the following, a brief introduction to \textbf{state space models} is provided.   The specific outline is given below:
\begin{enumerate}
    \item Modeling approaches for dynamical systems
    \item Formal definitions and change of bases on linear time invariant state space models.
    \item Discretization.
\end{enumerate}

\subsection{Modeling Approaches}
Mathematical models for dynamical systems have been studied for centuries. Two common approaches to modeling the relation between input and output of such systems are:

\begin{enumerate}
    \item \textbf{Input-output view}: In this approach, the system is directly characterized  by its input-output relation. The latter is typically described by means of a (high order) differential equation.

    \item \textbf{State-space view}: This approach introduces an auxiliary variable to model the system: the \textbf{state}. The latter provide a complete description of the configuration of the system at each time instant. As a consequence, the state at present time contains all the information about the past of the system that is relevant for its future evolution.
    The input-output view is still present, but now the state acts as a ``mediator'' between the input and the output.
\end{enumerate}

State space models are widely used in numerous fields, including mathematics, engineering, economics, and biology, and have recently been employed in the deep learning field to maintain a compressed representation of the context provided by previously seen tokens, where the tokens take on the role of the inputs. 

From now on, we focus only on state space models in which the state is defined over a finite-dimensional vector space.

\subsection{Definitions and changes of basis}
In this section, we aim to provide a brief introduction to linear state-space models. Both continuous- and discrete-time state-space models are defined along with a relation between the two.
We also define the important property of algebraic equivalence.

\begin{definition}
A {\em linear, continuous-time state space model} is a representation of a system in which the input $u(t)$ and the output $y(t)$, $t\in{\mathbb R}$, are continuous-time signals and their relation is given by:
\begin{equation}
\left\{
\begin{aligned}
    \dot{x}(t) &= A(t) x(t) + B(t) u(t) \\
    y(t) &= C(t) x(t)
\end{aligned}
\right.
\label{models:ct_ssm}
\end{equation}
where the continuous-time, vector-valued signal  $x(t) \in \mathbb{R}^{n}$ is called  {\em state} of the model.
The matrix $A(t) \in \mathbb{R}^{n \times n}$ is called  {\em state transition matrix},   $B(t) \in \mathbb{R}^{n \times m}$ is called  {\em input} matrix (or input to state matrix) and
$C(t) \in \mathbb{R}^{p \times n}$ is called {\em output} matrix (or state to output matrix).

A {\em linear, discrete-time state space model} is a representation of a system in which the input $u(k)$ and the output $y(k)$, $k\in{\mathbb Z}$, are discrete-time signals and their relation is given by:
\begin{equation}
\left\{
\begin{aligned}
    x(k) &= A(k) x(k-1) + B(k) u(k) \\
    y(k) &= C(k) x(k)
\end{aligned}
\right.
\label{models:dt_ssm}
\end{equation}
Again the discrete-time, vector-valued signal  $x(k) \in \mathbb{R}^{n}$ is called  {\em state} of the model.
Matrix dimensions and names are the same as those used for the model in (\ref{models:ct_ssm}).

\end{definition}

We refer to $n$ as the state space dimension, while $m$ and $p$ are referred to as the  input and output space dimensions, respectively. The first equation in (\ref{models:ct_ssm}) is a first order differential equation (or a set of first order differential equations for $n > 1$) describing how the state evolves over time. 
Once the state evolution is determined, the second equation in (\ref{models:ct_ssm}) yields the evolution of the variables of interest modeled by $y(t)$. 
Linear state-space models are often represented as triples in the form $\Sigma=(A(k),B(k),C(k))$ ($\Sigma=(A(t),B(t),C(t))$
for continuous-time models).
\\
The interpretation of the equations in (\ref{models:dt_ssm}) is with the difference that, given the discrete-time nature, the state evolution must obey a difference equation (or a set of difference equations).\\
If the matrices $A, B, C$ are independent of $t$ (or of $k$ in discrete-time), the model is said to be {\em time-invariant}.

There is an inherent redundancy in the state space description because we can arbitrarily select the basis in the state space $\mathbb{R}^{n}$ and this choice does not affect the input-output relation of the model. As a consequence, there are infinitely many state space models corresponding to the same input/output dynamics. 
Two linear models\footnote{Here we refer to discrete-time models but everything can be repeated {\em verbatim} for the continuous-time case.} $\Sigma_{1} = (A_1(k),B_1(k),C_1(k))$ and $\Sigma_{2} = (A_2(k),B_2(k),C_2(k))$ differing only for the choice of basis in the state space are said to be {\em algebraically equivalent}.
A simple and direct computation shows that, in this case,
there exists a non-singular matrix $T\in \mathbb{R}^{n\times n}$ such that
$A_2(k)=T^{-1}A_1(k)T$, $B_2(k)=T^{-1}B_1(k)$ and $C_2(k)=C_1(k)T$. The matrix $T$ induces the change of basis and is such that the states $x_1(t)$ of $\Sigma_1$ and
$x_2(t)$ of $\Sigma_2$ are related by $x_1(t)=Tx_2(t)$.

We can also perform a change of basis in the input or in the output space. Of course, these changes affect the input-output relationship. 
However, in the setting of this work, this is not a problem.
In fact, the input sequence is made up of tokens whose embedding in $\mathbb{R}^{m}$ is learned so that it can absorb the selection of the basis. The latter transforms the model by only changing $B(k)$ to $B(k)V$ where $V\in \mathbb{R}^{m\times m}$ is the (non-singular) matrix inducing the change of basis in the input space.
A similar argument holds for the change of basis in the output space which transforms the model by only changing $C(k)$ to $L^{-1}C(k)$ where $L\in \mathbb{R}^{p\times p}$ is the (non-singular) matrix inducing the change of basis in the output space.

\subsection{Discretization}
\label{sec:Appendix_disc}
Consider a continuous-time state space model 
$\Sigma_c = (A(t), B(t), C(t))$ whose input $u(t)$
is piece-wise constant on intervals of the same length $T$, so that for all $t_1,t_2\in[kT,(k+1)T)$, with
$k\in\mathbb{Z}$, $u(t_1)=u(t_2)$.
In this case, we can concentrate the input information into a new discrete-time signal $u_d(k):=u(kT)$ and by integrating 
the first of Equations (\ref{models:ct_ssm}) for each interval 
$[kT,(k+1)T)$, we obtain a discrete-time model where the discrete state and output are obtained by sampling the original state and output at the instants $kT$: $x_d(k):=x(kT)$,  $y_d(k):=y(kT)$.\\
In the case of a time-invariant continuous-time models, we can easily derive closed-form expressions for the discrete-time model $\Sigma_d = (A_d, B_d, C_d)$ which is time-invariant as well: 
\begin{equation}
\label{for:disc_ZOH}
\left\{
\begin{aligned}
    A_d &= e^{AT} \\
    B_d &= \int_{0}^{T} e^{A\tau}Bd\tau =A^{-1}\left(e^{AT} - I \right)B\\
    C_d &= C
\end{aligned}
\right.
\end{equation}
where the last equality in the expression of $B_d$ holds in the case of non-singular state matrix $A$.

\section{The S6 state space model}
\label{appendix:the_S6_model}
SSMs have recently been explored as efficient sequence models, motivated by their ability to capture long-range dependencies with favorable scaling, and as a structured alternative to attention \cite{SSMControlOverview,Strucutured_SSM,diagonal_SSM,S5_model,LRU}. Early work focused on linear time-invariant (LTI) models, which are computationally efficient but limited in expressivity: their fixed recurrence forces every token to influence the state, preventing the model from ignoring irrelevant inputs.  

The S6 model \cite{MAMBA} addresses this by introducing linear time-varying (LTV) dynamics. The state matrices $(A(k),B(k),C(k))$ depend on the input embedding, enabling token-dependent gating of information. This design, motivated by tasks such as \emph{induction head} and \emph{selective copy}, allows the model to update memory selectively.
Unlike LTI SSMs, the time-varying recurrence cannot be expressed as a convolution, precluding fast FFT-based implementations. To remain efficient, \cite{MAMBA} proposed a hardware-aware parallel algorithm that minimizes GPU memory traffic, enabling scalable training and inference.

To describe the model more concretely, consider its input representation. S6 operates on a multi-dimensional array $[B,L,D]$, with batch size $B$, sequence length $L$, and embedding dimension $D$. Each $L\times D$ matrix is an embedded sequence, where columns correspond to embedding features. The model processes these features independently by instantiating $D$ parallel SISO SSMs: the $i$-th SSM receives the $i$-th feature across all tokens. Each feature thus maintains its own state, while the recurrences are coupled through shared token-dependent matrices $B(t),C(t)$ and adaptive step sizes $\Delta_i(t)$, enabling selective memory updates.  

Formally, for each feature $i=1,\dots,D$, the model defines a SISO state space system:
\begin{equation*}
\Sigma^{(i)} =
\left\{
\begin{array}{l}
  \dot{x}(t)^{(i)} = A^{(i)} x(t)^{(i)} + B(t) u(t)_{i} \\
  y(t)^{(i)} = C(t) x(t)^{(i)}
\end{array}
\right.
\end{equation*}
where
\begin{itemize}
    \item $x(t)^{(i)}\in\mathbb{R}^n$ is the hidden state associated with feature $i$. The dimension $n$ is shared across features and treated as a tunable hyperparameter.  
    \item $u(t)_i$ denotes the $i$-th coordinate of the token embedding $u(t)\in\mathbb{R}^D$, which acts as the scalar input for the $i$-th SSM.  
    \item $A^{(i)}\in\mathbb{R}^{n\times n}$ is the transition matrix, chosen diagonal with learnable entries, and independently parameterized for each feature.  
    \item $B(t),C(t)\in\mathbb{R}^n$ are input- and output-mapping vectors, shared across features but dependent on the entire token embedding:
    \[
    B(t) = W_B u(t)^\top, \qquad C(t) = (W_C u(t)^\top)^\top ,
    \]
    with $W_B,W_C\in\mathbb{R}^{n\times D}$ learnable. This design makes the recurrence token-dependent, allowing the model to gate information based on content.  
    \item $y(t)^{(i)}$ represents the feature-level output at time $t$.  
\end{itemize}

Note that this is a continuous-time formulation. In practice, the typical sequence inputs, such as text, are inherently discrete. A standard way to bridge this gap is to view $u(t)$ as a piecewise constant signal derived from the discrete embeddings, and then apply a discretization procedure. In S6, the chosen approach is zero-order hold (ZOH), which admits closed-form discrete-time matrices (see Sec.~\ref{sec:Appendix_disc}).

In the S6 model, discretization is applied independently to each feature-specific system $\Sigma^{(i)}$. Importantly, the sampling interval is not fixed but \textbf{input-dependent}: for each token $u(t)$, the model defines a vector of adaptive step sizes
\begin{equation}
\label{for:vector_Delta}
\Delta(t) = \softplus(W_D u(t)^\top) \in \mathbb{R}^D ,
\end{equation}
where $W_D\in\mathbb{R}^{D\times D}$ is learnable and the softplus $\softplus(\cdot)$ ensures positivity. Each $\Delta_i(t)$ controls the discretization of the $i$-th feature’s SSM. Rather than being literally interpreted as different sampling times, these values can be understood as feature-specific gates that balance the influence of the past state and the current input when forming the update.  

The resulting discrete-time system for feature $i$ is
\begin{equation}
\label{for:model_not_expanded}
\Sigma^{(i)} =
\left\{
\begin{array}{l}
  x(k)^{(i)} = e^{A^{(i)}\Delta_i(k)} x(k-1)^{(i)} + \big(A^{(i)}\big)^{-1} \big(e^{A^{(i)}\Delta_i(k)}-I \big) B(k) u(k)_{i} \\
  y(k)^{(i)} = C(k) x(k)^{(i)}
\end{array}
\right.
\end{equation}

In S6, $A^{(i)}$ is chosen diagonal with entries $\lambda_j^{(i)}$ and the previous expression simplifies to
\begin{equation}
\label{for:mamba_model_expanded}
\Sigma^{(i)} =
\left\{
\begin{array}{l}
  x(k)^{(i)} =  
    \begin{bmatrix}
    e^{\lambda_{0}^{(i)}\Delta_i(k)} &  &  \\
                & \ddots & \\
                &         & e^{\lambda_{n-1}^{(i)}\Delta_i(k)}
    \end{bmatrix}
  x(k-1)^{(i)} +
    \begin{bmatrix}
    \frac{(e^{\lambda_{0}^{(i)}\Delta_i(k)}-1)B_0(k)}{\lambda_{0}^{(i)}}\\
    \vdots            \\
    \frac{(e^{\lambda_{n-1}^{(i)}\Delta_i(k)}-1)B_{n-1}(k)}{\lambda_{n-1}^{(i)}}
    \end{bmatrix}
  u(k)_{i} \\
  y(k)^{(i)} =
    \begin{bmatrix}
    C_0(k) & \cdots & C_{n-1}(k)
    \end{bmatrix}
  x(k)^{(i)}
\end{array}
\right.
\end{equation}
The S6 parameterization is designed to enable input-dependent selectivity. The input and output mappings are token-dependent,
\[
B(k) = W_B u(k)^\top, \qquad C(k) = (W_C u(k)^\top)^\top ,
\]  
which allows the model to decide at each step whether to admit information into the state or propagate it to the output \cite{MAMBA}.  

Selectivity is further controlled by the adaptive step sizes : 
\[
\Delta(k) = \softplus(W_D u(k)^\top), \qquad k\in\mathbb{Z}.
\]  
In the discrete update (Eq.~(\ref{for:mamba_model_expanded}))  if the eigenvalues $\lambda^{(i)}_{j}$ are in the left half of the complex plane (that is, they correspond to stable modes in continuous time), $\Delta_i(k)$ interpolates between two extremes:  
\begin{itemize}
    \item $\Delta_i(k)\to\infty$: past context is forgotten, the state depends only on $u(k)_i$;  
    \item $\Delta_i(k)\to 0$: the input is ignored, the state is preserved.  
\end{itemize}  

This mechanism provides fine-grained control over memory and update. 

Finally, stability is ensured by constraining the eigenvalues of $A^{(i)}$ to be negative via  
\[
\lambda_j^{(i)} = -e^{\mu_j^{(i)}} ,
\]  
with $\mu_j^{(i)}$ unconstrained.

\section{Extended Derivations for COFFEE}\label{appendix:derivation_of_our_model}

Our formulation builds on the S6 model (Appendix \ref{appendix:the_S6_model}). 
The derivation proceeds in three steps:  
\begin{enumerate}
    \item Linearize the exponential in Equation (\ref{for:model_not_expanded}) via a first-order Taylor expansion to clarify the role of $\Delta(t)$ and to enhance the mechanistic interpretability.
    \item Incorporate state feedback to enable context-dependent selectivity. 
    \item Remove parameter redundancy through changes of basis in the state and input spaces.
\end{enumerate}

\subsection{Simplified Dynamics via Linearization}

Expanding the exponential in Equation (\ref{for:model_not_expanded}) around $\Delta_i(k)=0$ gives
\[
\Sigma^{(i)} =
\begin{cases}
  x(k)^{(i)} = \big(I + A^{(i)}\Delta_i(k)\big)x(k-1)^{(i)} 
  + \big(A^{(i)}\big)^{-1}\!\left(I + A^{(i)}\Delta_i(k) - I\right)B(k)u(k)_i, \\[0.5ex]
  y(k)^{(i)} = C(k)x(k)^{(i)} ,
\end{cases}
\]
which simplifies to
\begin{equation}
\label{for:linearized_model}
\Sigma^{(i)} =
\begin{cases}
  x(k)^{(i)} = \big(I + A^{(i)}\Delta_i(k)\big)x(k-1)^{(i)} + \Delta_i(k)B(k)u(k)_i, \\[0.5ex]
  y(k)^{(i)} = C(k)x(k)^{(i)} .
\end{cases}
\end{equation}

In S6, the matrices $B(k), C(k)$ are shared across all subsystems and depend on the input embedding. However, after discretization, the effective input-to-state mapping:
\[
\big(A^{(i)}\big)^{-1}\!\big(e^{A^{(i)}\Delta_i(k)} - I\big)B(k),
\]
involves $A^{(i)}$, making it inherently feature-specific. This motivates assigning each subsystem its own time-invariant matrix $B^{(i)}$. We additionally assign feature-specific $C^{(i)}$ to maintain symmetry in the model. We therefore equip each subsystem with feature-specific $B^{(i)}$ and $C^{(i)}$, both time-invariant.\footnote{Time variation arises solely through $\Delta_i(k)$.} This yields:
\begin{equation}
\label{for:our_model_only_lin}
\Sigma^{(i)} =
\begin{cases}
  x(k)^{(i)} = \big(I + A^{(i)}\Delta_i(k)\big)x(k-1)^{(i)} + \Delta_i(k)B^{(i)}u(k)_i \\
  y(k)^{(i)} = C^{(i)}x(k)^{(i)} ,
\end{cases}
\end{equation}
where $B^{(i)} \in \mathbb{R}^n$ is a column vector, $C^{(i)} \in \mathbb{R}^n$ a row vector, and $A^{(i)} \in \mathbb{R}^{n\times n}$ diagonal. Since (\ref{for:our_model_only_lin}) is derived via Taylor expansion, the model is accurate if the $\Delta_i(k)$ remain small. Following S6, we set
\[
\Delta(k) = \sigmoid(W_D u(k)^\top), \quad W_D \in \mathbb{R}^{D \times D}.
\]

\paragraph{Parameter count.}  
The resulting model has $3nD + D^2$ parameters:
\begin{itemize}
    \item $nD$ from $\{B^{(i)}\}$,  
    \item $nD$ from $\{C^{(i)}\}$,  
    \item $nD$ from the diagonals of $\{A^{(i)}\}$,  
    \item $D^2$ from $W_D$.  
\end{itemize}
This matches S6, which has $nD$ parameters in each of $W_B$ and $W_C$, $nD$ for $\{A^{(i)}\}$, and $D^2$ for $W_D$, again totaling $3nD + D^2$. We remark that these counts refer only to the SSM block. The full architecture (presented in Appendix \ref{appendix:core_module_architecture}) includes an additional $|M|D$ parameters from the embedding matrix, for a total of $3nD + D^2 + |M|D$, where $M$ is the vocabulary size.

\subsection{Feedback for Context-Based Selectivity}

Our model introduces context-based selectivity by letting the update gates depend on the hidden state, rather than the raw input token. In S6, gating is computed as:
\[
\Delta(k) = \zeta(W_D u(k)^\top), \qquad W_D \in \mathbb{R}^{D\times D}.
\] 
where we recall that $\zeta(\cdot)$ is the softplus function.
In contrast, we replace this token-based rule with a state-feedback law:
\begin{equation}
\label{eq:Delta-feedback-appendix}
\Delta_i(k) = \sigmoid\big(w_D^{(i)} \odot x(k-1)^{(i)}\big), \qquad w_D^{(i)} \in \mathbb{R}^n,
\end{equation}
where gating is conditioned on the compressed context encoded in the hidden state of each subsystem $\Sigma^{(i)}$. In addition to introducing context-dependent selectivity, this change reduces the gating parameters from $D^2$ in S6 to $Dn$. Since typically $n \ll D$, this corresponds to a considerable reduction in the number of gating parameters.
The state-dependent gating \(\Delta_i(k)\) is incorporated into the subsystem dynamics (\ref{for:our_model_only_lin}) as
\begin{equation}
\label{eq:model_with_feedback_app}
\Sigma^{(i)} =
\begin{cases}
  x(k)^{(i)} =
    (I+A^{(i)}\mathrm{diag}(\Delta_i(k)))\,x(k-1)^{(i)} 
    + \mathrm{diag}(\Delta_i(k)) B^{(i)}\,u(k)_i, \\
  y(k)^{(i)} = C^{(i)} x(k)^{(i)}.
\end{cases}
\end{equation}

\subsection{Eliminating Redundancy via Change of Basis}\label{proof-prop-changebasis}

The number of parameters in (\ref{eq:model_with_feedback_app}) can be reduced through a suitable change of basis. This is formalized in Proposition \ref{prop:redundancy_elimination} in the main text; here we provide the proof.

{\bf Proof of Proposition \ref{prop:redundancy_elimination}.} Let $j$ be the index of the embedding vector $v^{(j)}$ whose entries $v_{i}^{(j)}$ are all nonzero, and consider the family of systems in Equation (\ref{eq:model_with_feedback_app}).
We rewrite this family as
\begin{equation}
\label{eq:linearized_model-newbasisinput}
\begin{cases}
  x(k)^{(i)} = \big(I + A^{(i)}\mathrm{diag}(\Delta_i(k)))\big)\,x(k-1)^{(i)} 
              + \mathrm{diag}(\Delta_i(k))) [B^{(i)} v_{i}^{(j)}][u(k)_{i}/v_{i}^{(j)}], \\
  y(k)^{(i)} = C^{(i)} x(k)^{(i)} .
\end{cases}
\end{equation}
It is therefore apparent that the $j$-th embedding vector can be normalized to $\mathbf{1}$
a the price of rescaling each $B^{(i)}$ by multiplying each of its entries by $v_{i}^{(j)}$.
This is indeed the simplest example of change of basis in the input space which is simply $\mathbb{R}$.

Now we are ready for a change of basis on the state space which does not affect the input/output behavior of the state model.
We select the change of basis induced by the matrix
$T_i:={\rm diag}\big(B^{(i)}_0v_{i}^{(j)},\dots, B^{(i)}_{n-1}v_{i}^{(j)}\big)$, with $B^{(i)}_k$ being the $k$-th entry of the vector $B^{(i)}$.
This transformation does not affect the state transition matrix but normalizes the input matrix to $\mathrm{diag}(\Delta_i(k)))\mathbf{1}=\Delta_i(k)$. Finally, after the change of basis, the output matrix becomes $(B^{(i)})^\top v_{i}^{(j)}\odot C^{(i)}$, where $\odot$ denotes the Hadamard product. \qed

Notice that assuming that each entry of $B^{(i)}$ is non-zero is very reasonable as if one entry, say the $k$-th, is zero it means that the $k$-th component of the state is not affected by the input and hence it can be eliminated without affecting the input/output relationship of the model. The fact that at least one of the embeddings has all entries which are non-zero is a very reasonable assumption, as the embedding parameters are learned.

In summary, the change-of-basis argument reveals two sources of redundancy:\\
(i) One embedding vector can be fixed arbitrarily (e.g., to $\mathbf{1}$) without altering the input-output behavior, which reduces the embedding parameters from $|M|D$ to $(|M|-1)D$;\\ 
(ii) $B^{(i)}$ and $C^{(i)}$ need not be learned independently: any rescaling of $B^{(i)}$ can be absorbed into $C^{(i)}$, so only their product matters. One set can therefore be fixed, saving $nD$ parameters.

After applying all the steps described above, the resulting COFFEE dynamics is given in Equation (\ref{eq:final_model}) of the main text. The overall parameter count decreases from $3nD + D^2 + |M|D$ in S6 to $3nD + (|M|-1)D$ in our model.

\section{Parallelization}
\label{Appendix:parallelization}

In this section, we provide details on the parallelization technique adopted to efficiently parallelize our model. The following exposition is based on \cite{DEER_lim} and \cite{gonzalez2024towards}.  

Let $x(k) \in \mathbb{R}^{\tilde{n}}$\footnote{$\tilde{n}$ may denote the dimension of each $x(k)^{(i)}$ in Equation \ref{eq:final_model} (in which case $\tilde{n}=n$) or the whole state obtained by stacking the $x(k)^{(i)}$ on top of each others (in which case $\tilde{n}=Dn$).} denote the state of a nonlinear Markovian discrete-time state-space model, with nonlinear transition dynamics governed by a function $f: \mathbb{R}^{\tilde{n}} \rightarrow \mathbb{R}^{\tilde{n}}$. We denote by $x_{1:L}$ the collection of states $x(1), \dots, x(L)$, where $L$ is the sequence length.  

Our goal is to find the unique trajectory $x^{*}(1), \dots, x^{*}(L)$ such that:
\[
\begin{cases}
x^{*}(1) = f(x(0)) \\
x^{*}(2) = f(x^{*}(1)) \\
\vdots\\
x^{*}(L) = f(x^{*}(L-1)) 
\end{cases}
\]
If we define the so-called residual as:
\begin{equation*}
    r(x_{1:L}) \;:=\;  [x(1)-f(x(0)),x(2)-f(x(1)),...,x(L)-f(x(L-1))] \in \mathbb{R}^{L\tilde{n}}
\end{equation*}
mathematically, we can define a vector-valued function $r: \mathbb{R}^{L\tilde{n}} \rightarrow \mathbb{R}^{L\tilde{n}}$, and our objective is to find the zero of this function. One approach to solve this problem is to apply Newton's method.  

We begin by selecting an initial guess $x_{1:L}^{(0)}$, representing a tentative trajectory of the nonlinear state-space model at the zeroth iteration. Then, we iteratively update this guess until convergence by performing the following steps:
\begin{enumerate}
    \item Given the $(i-1)$-th solution $x^{(i-1)}_{1:L}$ perform a first order Taylor expansion of $r(\cdot)$ centered at $x^{(i-1)}_{1:L}$
    \item From the previous point we obtain $r(x_{1:L}) = r(x^{(i-1)}_{1:L}) + J_r(x^{(i-1)}_{1:L})(x_{1:L} - x^{(i-1)}_{1:L}) + o(\cdot)$
    \item Solve $0 = r(x^{(i-1)}_{1:L}) + J_r(x^{(i-1)}_{1:L})(x_{1:L} - x^{(i-1)}_{1:L})$ in $x_{1:L}$
    \item Set the solution to previous equation as $x^{(i)}_{1:L}$
\end{enumerate}
Concretely, each iteration involves solving the following linear system:
\begin{equation*}
    J_r(x^{(i-1)}_{1:L})(x_{1:L} - x^{(i-1)}_{1:L}) = -r(x^{(i-1)}_{1:L})
\end{equation*}
By evaluating $J_r(x^{(i-1)}_{1:L})$ we obtain:
\[
\begin{bmatrix}
I_n & 0 & 0 & \cdots & 0 & 0 \\
-\dfrac{\partial f}{\partial x}\!\bigl(x^{(i-1)}(1)\bigr) & I_n & 0 & \cdots & 0 & 0 \\
0 & -\dfrac{\partial f}{\partial x}\!\bigl(x^{(i-1)}(2)\bigr) & I_n & \cdots & 0 & 0 \\
\vdots & \vdots & \ddots & \ddots & \vdots & \vdots \\
0 & 0 & \cdots & -\dfrac{\partial f}{\partial x}\!\bigl(x^{(i-1)}(L-2)\bigr) & I_n & 0 \\
0 & 0 & \cdots & 0 & -\dfrac{\partial f}{\partial x}\!\bigl(x^{(i-1)}(L-1)\bigr) & I_n
\end{bmatrix}
\]
which is a block bidiagonal matrix in $\mathbb{R}^{L\tilde{n} \times L\tilde{n}}$. Although this matrix is invertible, directly solving the linear system by computing $J_r(x^{(i-1)}_{1:L})^{-1}$ is computationally intractable at large scale.  

However, we can exploit the block bidiagonal structure to efficiently solve the system. In particular, by defining
\[
\Delta x^{(i)}(k) \;:=\; x^{(i)}(k) - x^{(i-1)}(k)
\]
and expanding the equations accordingly, we can avoid the full matrix inversion while still performing the Newton update. In fact, by forward substitution on:
\begin{equation*}
    J_r(x^{(i-1)}_{1:L})(x_{1:L} - x^{(i-1)}_{1:L}) = -r(x^{(i-1)}_{1:L})
\end{equation*}
we obtain the following first order linear recurrence:
\begin{equation*}
\begin{cases}
    \Delta x^{(i)}(1) = - r_1(x^{(i-1)}_{1:L})\\
    \Delta x^{(i)}(k) = \dfrac{\partial f}{\partial x}(x^{(i-1)}(k-1)) \Delta x^{(i)}(k-1) - r_k(x^{(i-1)}_{1:L})
\end{cases}
\end{equation*}
where 
\[
r_k(x^{(i-1)}_{1:L}) \;:=\; x^{(i-1)}(k) - f(x^{(i-1)}(k-1)), \quad k = 1, \dots, L.
\]  
To solve this first-order linear recurrence, we employ Blelloch's parallel scan algorithm \cite{blelloch1990prefix}.

According to \cite{blelloch1990prefix}, we need to store the pairs 
\[
e_k \;:=\; \Bigg(\frac{\partial f}{\partial x}(x^{(i-1)}(k)), \; -r_k(x^{(i-1)}_{1:L}) \Bigg).
\]  
Since $\frac{\partial f}{\partial x}(x^{(i-1)}(k))$ is generally a dense matrix in $\mathbb{R}^{\tilde{n} \times \tilde{n}}$ and there are $L$ such matrices to store, the memory requirement scales as $\mathcal{O}(L\tilde{n}^2)$. Furthermore, combining the various $e_k$ according to Blelloch's algorithm requires $L$ matrix multiplications of dense $\tilde{n} \times \tilde{n}$ matrices, resulting in a total computational cost of $\mathcal{O}(L\tilde{n}^3)$.  

To mitigate these computational and memory demands, \cite{gonzalez2024towards} proposed a modification in which only the diagonals of the Jacobians are considered. In this case, $\frac{\partial f}{\partial x}(x^{(i-1)}(k))$ becomes a diagonal matrix in $\mathbb{R}^{\tilde{n} \times \tilde{n}}$. Consequently, storing the pairs 
\[
e_k \;:=\; \Bigg(\frac{\partial f}{\partial x}(x^{(i-1)}(k)), \; -r_k(x^{(i-1)}_{1:L}) \Bigg)
\]  
now requires only $\mathcal{O}(L\tilde{n})$ memory. Moreover, multiplying diagonal matrices reduces to performing the Hadamard product of their diagonals, which has complexity $\mathcal{O}(\tilde{n})$. Thus, the $L$ dense matrix multiplications are replaced by $L$ Hadamard products of $\tilde{n}$-dimensional vectors, yielding a total computational complexity of $\mathcal{O}(L\tilde{n})$.  

This modified algorithm is referred to as \emph{quasi-DEER} in \cite{gonzalez2024towards}.
\newline
\newline
Another contribution of \cite{gonzalez2024towards} concerns the improved numerical stability of the algorithm. The key idea is to reformulate the trajectory evaluation problem as an optimization problem. By doing so, it becomes evident that the previously described solution corresponds to a Gauss-Newton method. Introducing a trust region for the update steps transforms this into the so-called Levenberg-Marquardt algorithm.  

The optimization problem obtained by applying the Levenberg-Marquardt algorithm is convex and satisfies Slater's constraint qualification, which guarantees strong duality. The approach then consists of treating the unique Lagrange multiplier as a hyperparameter of the model and minimizing only the associated Lagrangian. When minimizing the Lagrangian with fixed Lagrange multipliers, one can observe, as noted in \cite{sarkka2020levenberg}, a connection to the MAP estimate of $x_{1:L}$ in a linear Gaussian state-space model. This estimate can be computed via a Kalman smoother.  

In this context, the Kalman smoother can be parallelized as described in \cite{sarkka2020temporal}. Specifically, by considering a set $\mathcal{F}$ consisting of appropriate conditional densities and likelihoods, and by defining a suitable associative operator $\otimes$, it can be shown that $(\mathcal{F}, \otimes)$ forms a monoid. This structure allows, once again, to leverage Blelloch's parallel scan algorithm to evaluate the Kalman smoother in $\mathcal{O}(\log L)$ steps on a parallel machine.  

In \cite{gonzalez2024towards}, this algorithm is referred to as \textbf{E}valuating \textbf{L}evenberg-Marquardt with \textbf{K}alman (ELK).

\section{Induction Head Task}
\label{appendix:IH_task}

The term ``induction heads'' is used by \cite{InductionHead} to denote a type of attention heads in the Transformers architecture. These heads
learn the ability to complete a sequence by memorizing the symbol (or symbols) following a certain subsequence called ``trigger'' and by producing such a symbol whenever the trigger appears again.
In \cite{InductionHead}, strong evidence is presented that supports the assumption that this ability might be one of the fundamental mechanisms for "in-context learning".
For this reason, such an ability is a benchmark to assess to what extent the models are capable of identifying patterns within sequences.

\subsection{Task description}

Consider a sequence of symbols extracted from a certain finite vocabulary $V$.
In our setting, we select 
\begin{equation*}
    V = \{1,2,3,4,5,6,7\}
\end{equation*}

To test the model's ability to capture patterns, the model is presented with a sequence of arbitrary length that follows the structure outlined below:
\begin{equation}
    \label{task:sequence_structure}
    \text{noise 1}\ \| \ \text{trigger}\ \| \ \text{target}\ \| \ \text{noise 2}\ \| \ \text{trigger}
\end{equation}
where $\|\ $ denotes sequence concatenation, and noise 1, noise 2, trigger, and target indicate subsequences of symbols. Each sequence is constructed in such a way that the same trigger subsequence appears exactly twice. The two noise subsequences may contain the target subsequence, but they must not include the trigger subsequence. The trigger subsequence serves to indicate the beginning of a pattern to be reproduced. Upon encountering the trigger subsequence, the model is expected to recognize that it must ``pay attention" to the subsequent elements, called {\em target} subsequence, and retain this information.
While the trigger subsequence is fixed, the target subsequence varies across sequences. We consider the task successfully completed when the model outputs the target subsequence after observing the final trigger subsequence in Equation (\ref{task:sequence_structure}). This outcome would demonstrate that the model has correctly learned the pattern ``$\text{trigger } \| \ \text{target}$".
\newline
\newline
When the sequence length is fixed, the following steps are necessary to build an appropriate sequence:
\begin{enumerate}
    \item Choose the length of the trigger subsequence
    \item Choose a fixed trigger subsequence with symbols from $V$
    \item Choose and fix the length of the target subsequence
\end{enumerate}

\begin{remark}
Fixed trigger subsequence means that once it is selected (e.g. randomly), it must be used consistently to generate all training, validation, and test sequences. In other words, in all training, validation and test sequences, the trigger subsequence is always the same in both of its occurrences in Equation (\ref{task:sequence_structure}).    
\end{remark}

Let us denote by $L_{\text{seq}}$, $L_{\text{tri}}$, and $L_{\text{tar}}$ the lengths of the sequence, trigger, and target subsequences, respectively. These lengths should be chosen such that:
\begin{equation*}
    L_{\text{seq}}- 2L_{\text{tri}} - L_{\text{tar}} \geq 1
\end{equation*}
to ensure that there is at least one of the two ``noise'' subsequences, consisting of at least one element.   In our experiments, we considered sequences constructed according to the following combinations:
\begin{enumerate}
    \item \textbf{Trigger} and \textbf{target} subsequences both of length one
    \item \textbf{Trigger} subsequence of length two and \textbf{target} of length one
    \item \textbf{Target} subsequence of length two and \textbf{trigger} of length one
\end{enumerate}
Once the trigger and the target length are fixed, to generate admissible sequences we randomly select the position of the initial symbol of the first ``trigger'' subsequence, the target subsequence and the two noise subsequences.
We use suitable uniform discrete independent random variables for all these selections with the caution of discarding sequences in which at least one of the target and the noise subsequences contains the trigger.
\newline
\newline
\begin{remark}
Whenever the length of the target subsequence is greater than $1$, extra caution is required. In fact, in this case,
when the model encounters the last trigger subsequence, it can output only one symbol and cannot complete the target subsequence. To address this issue, we pad the input sequence by concatenating it with the padding sequence made of  \(L_{\text{tar}} - 1\) copies of the special symbol ``0''.  For example, an admissible sequence with \(L_{\text{tri}} = 3\) and \(L_{\text{tar}} = 3\) could be:
    \[
    \underbrace{3,2,6,}_{\text{noise}} \underbrace{5,6,7}_{\text{trigger}} \underbrace{2,4,3,}_{\text{target}} 
    \underbrace{1,2,2,6,}_{\text{noise}}
    \underbrace{5,6,7,}_{\text{trigger}}
    \underbrace{0,0}_{\text{padding}}
    \]
In this way, the model can observe the entire trigger subsequence and produce the complete target subsequence. Upon encountering the last symbol, ``7", the correct first output would be ``2", and the subsequent zeros would allow the model to also output the rest of the target subsequence, i.e.,``4" and ``3".
\end{remark}
 
All the generated input sequences are transformed, by the usual embedding process, into sequences of vectors. The latter are used as inputs to a state-space model. Specifically, given a linear discrete-time state-space model
\[
\left\{
\begin{aligned}
    x(k) &= A x(k-1) + B u(k) \\
    y(k) &= C x(k) + D u(k)
\end{aligned}
\right.
\]
embedding vectors of symbols are the various $u(k)$ for $k=0,...,L_{\text{seq}}-1$. For example, given the sequence:
\[
\underbrace{3,2,6,}_{\text{noise}} \underbrace{5,6,7}_{\text{trigger}} \underbrace{2,4,3,}_{\text{target}} 
\underbrace{1,2,2,6,}_{\text{noise}}
\underbrace{5,6,7,}_{\text{trigger}}
\underbrace{0,0}_{\text{padding}}
\]
$u(0)$ is the embedding of the symbol ``3", $u(1)$ is the embedding of the symbol ``2", and so on.

We finally emphasize that this task can be defined over any set $V$. In our case, we used $V = \{1,2,3,4,5,6,7\}$, but one could choose a larger set or a vocabulary consisting of different symbols, such as $V=\{\text{``cat"},\text{``dog"}, ...\}$. The elements of the vocabulary are symbols, and the relevant thing is to highlight their mutual relations.

\section{Core Module Architecture}
\label{appendix:core_module_architecture}
We now provide a description of the core module architecture employed to address the induction head task. A detailed account of all steps involved in data processing is  presented. The operations performed by the core module in our setup are as follows:
\begin{enumerate}
    \item The core module receives a sequence or a batch of sequences as input. Each sequence is built according to the rules presented in Appendix \ref{appendix:IH_task}. The input can be regarded as an array of dimension [$B$, $L$], where $B$ denotes the batch size and $L$ the length of each sequence within the batch. \footnote{More generally, we use the notation [$d_1$,..., $d_l$] to indicate an array having $l$ dimensions, where the first dimension is $d_1$ and the $l$-th dimension is $d_l$}
 
    \item Each symbol is mapped to a real-valued vector, referred to as the  {\em embedding vector}. The dimensionality of the embedding space is a user-defined hyperparameter and can be adjusted according to the specific requirements of the task.
    Formally, the embedding procedure consists in applying the following \textbf{learned} map:
    \[
    \textrm{emb}: M \rightarrow \mathbb{R}^{D}
    \]
    where $M$ is the model's vocabulary while $D$ denotes the dimension of the embedding  space and sometimes is also called  {\em model dimension}.
    \newline
    \begin{remark}
        Notice that, the domain of the $\textrm{emb}$ map is $M$. Since we assume that $V \subseteq M$ ($V$ is the vocabulary used to build sequences), this transformation is well-defined: it guarantees that every symbol in any input sequence has a corresponding embedding vector.
    \end{remark}

    After applying this mapping, we obtain a three-dimensional array with shape [$B$, $L$, $D$], where $D$ denotes the embedding dimensionality. 
   
    \item The resulting sequence of embedding vectors is fed into a state space model, which contains additional learnable parameters.
    
    \item For each input (i.e. the embedding vector of a symbol) the model outputs another vector of the same dimension, specifically a vector in $\mathbb{R}^{D}$.
    After processing all input sequences, the state space model produces a multidimensional array of shape [$B$, $L$, $D$]. For each batch index $b = 0, \ldots, B-1$, this corresponds to an $L \times D$ matrix containing all output vectors associated with the input sequence.

    \item In general, the outputs produced by the state space model do not correspond exactly to the embedding vectors associated with symbols in $M$. However, the ultimate objective is for the model to generate symbols from the set $M$. To achieve this, the first step involves computing the distance between each output vector and every embedding vector corresponding to symbols in $M$. This operation results in a multidimensional array of shape [$B$, $L$, $|M|$].
  
    After having evaluated the distances we have:
    \[\mathrm{Distances} = 
    \begin{bmatrix}
        \begin{bmatrix}
            d_{0,0}^{(0)} & \cdots & d_{0,|M|-1}^{(0)}\\
            \vdots & \ddots  & \vdots \\
            d_{L-1,0}^{(0)} & \cdots & d_{L-1,|M|-1}^{(0)}\\
        \end{bmatrix}\\
        \vdots \\
        \begin{bmatrix}
            d_{0,0}^{(B-1)} & \cdots & d_{0,|M|-1}^{(B-1)}\\
            \vdots & \ddots  & \vdots \\
            d_{L-1,0}^{(B-1)} & \cdots & d_{L-1,|M|-1}^{(B-1)}\\
        \end{bmatrix}
    \end{bmatrix}
    \]
    where $d_{l,m}^{(b)}$ denotes the distance between the output vector at time $l$ of batch $b$ and the embedding vector of the symbol $m \in M$, for every $l = 0, \ldots, L-1$, $b = 0, \ldots, B-1$, and $m \in M$.
    \item We now apply the softmin\footnote{Given a vector $x = (x_1, x_2, \dots, x_n) \in \mathbb{R}^n$, the softmin function produces another vector $\softmin(x)\in \mathbb{R}^n$ whose entries are defined by 
    \[
    \softmin(x)_i = \frac{e^{-x_i}}{\sum_{j=1}^n e^{-x_j}}, \quad  i = 1, \dots, n.
    \]} function along the rows of the matrices within the multidimensional array $\mathrm{Distances}$. This operation produces another multidimensional array, denoted as $\mathrm{Softmin}$, which has the same dimensions as $\mathrm{Distances}$, namely [$B$, $L$, $|M|$].

    By applying the softmin function to the rows of $\mathrm{Distances}$ array, we obtain a probability  distribution where the smallest $d_{l,m}^{(b)}$ for every $l=0,...,L-1$, $b=0,...,B-1$ and $m \in M$ has the maximum probability. 
    \newline
    \newline
    The resulting matrix after this step is:
    \[\mathrm{Softmin} = 
    \begin{bmatrix}
        \begin{bmatrix}
            s_{0,0}^{(0)} & \cdots & s_{0,|M|-1}^{(0)}\\
            \vdots & \ddots  & \vdots \\
            s_{L-1,0}^{(0)} & \cdots & s_{L-1,|M|-1}^{(0)}\\
        \end{bmatrix}\\
        \vdots \\
        \begin{bmatrix}
            s_{0,0}^{(B-1)} & \cdots & s_{0,|M|-1}^{(B-1)}\\
            \vdots & \ddots  & \vdots \\
            s_{L-1,0}^{(B-1)} & \cdots & s_{L-1,|M|-1}^{(B-1)}\\
        \end{bmatrix}
    \end{bmatrix}
    \]
    
    \item After applying the softmin function, we then apply an element-wise transformation known as the  logit\footnote{The logit function is defined as the inverse of the sigmoid function.  
    For a real number \( p \in (0,1) \), it is given by:
    \[
    \logit(p) = \log\left( \frac{p}{1 - p} \right)
    \]}  function. The logit function maps probabilities from the interval $(0,1)$ to the entire real line $\mathbb{R}$.

    In this manner, we obtain another multidimensional array, which we denote as $\mathrm{Logits}$. Each entry is given by $\mathrm{Logits}_{k,i,j} = \logit(\mathrm{Softmin}_{k,i,j})$. Naturally, the $\mathrm{Logits}$ array has the same dimensions as $\mathrm{Softmin}$, namely [$B$, $L$, $|M|$]. The resulting matrix after this step is:
    \[\mathrm{Logits} = 
    \begin{bmatrix}
        \begin{bmatrix}
            \logit_{0,0}^{(0)} & \cdots & \logit_{0,|M|-1}^{(0)}\\
            \vdots & \ddots  & \vdots \\
            \logit_{L-1,0}^{(0)} & \cdots & \logit_{L-1,|M|-1}^{(0)}\\
        \end{bmatrix}\\
        \vdots \\
        \begin{bmatrix}
            \logit_{0,0}^{(B-1)} & \cdots & \logit_{0,|M|-1}^{(B-1)}\\
            \vdots & \ddots  & \vdots \\
            \logit_{L-1,0}^{(B-1)} & \cdots & \logit_{L-1,|M|-1}^{(B-1)}\\
        \end{bmatrix}
    \end{bmatrix}
    \]
    \item To conclude, the core module architecture produces as output two multidimensional arrays. The first is exactly $\mathrm{Logits}$. The second is a multidimensional array of dimensions [$B$, $L$], obtained by applying the $\argmax$ along the last dimension of $\mathrm{Logits}$. This means that we compute the $\argmax$ over all rows of every matrix within the multidimensional array $\mathrm{Logits}$. The resulting array of shape [$B$, $L$] is denoted as $\mathrm{Predictions}$. Formally, this matrix is constructed as follows:
    \[
    \mathrm{Predictions} = 
    \begin{bmatrix}
        \argmax_{\substack{m \in M}} \quad \logit_{0,m}^{(0)} & \cdots & \argmax_{\substack{m \in M}} \quad \logit_{L-1,m}^{(0)}\\
        & \vdots & \\
        \argmax_{\substack{m \in M}} \quad \logit_{0,m}^{(B-1)} & \cdots & \argmax_{\substack{m \in M}} \quad \logit_{L-1,m}^{(B-1)}
    \end{bmatrix}
    \]
    This matrix, for each row (i.e. for each batch), contains the predicted symbols for all time steps $l = 0, \ldots, L-1$. Notice that, as a consequence of the previous operations, taking the $\argmax$ corresponds to selecting, for every batch and every time step, the symbol whose embedding vector is closest (with respect to the Euclidean norm) to the output produced by the state space model.
    \newline
    \begin{remark}
        The multidimensional array $\mathrm{Predictions}$ can be obtained in multiple ways. For each $l = 0, \dots, L-1$, the predicted symbol $m_{\text{pred}} \in M$ is the one closest to the output vector of the state-space model. Therefore, taking the $\argmin$ over the rows of the matrices in the $\mathrm{Distances}$ array, or the $\argmax$ over the rows of the matrices in the $\mathrm{Softmin}$ or $\mathrm{Logits}$ arrays, will yield the same $\mathrm{Predictions}$ matrix.    
    \end{remark}
    We use cross-entropy loss as the loss function, which requires the logits as input. Therefore, the $\mathrm{Logits}$ array is used for computing the loss, while the $\mathrm{Predictions}$ array is used to evaluate the accuracy.
\end{enumerate}
The core module architecture is summarized in the following block diagram:
\begin{center}
\begin{tikzpicture}[
  node distance=1.5cm and 2.5cm,
  block/.style={rectangle, draw, minimum width=3cm, minimum height=1cm, align=center},
  arrow/.style={-{Latex}, thick},
  scale=0.75,every node/.style={scale=0.75},
]

\node[block] (sequence) {Input sequence};
\node[block, below=0.5cm of sequence] (embedding) {Embedding};
\node[block, below=0.5cm of embedding] (ssm) {State space model};
\node[block, below=0.5cm of ssm] (distance) {Distances evaluation};
\node[block, below=0.5cm of distance] (softmin) {softmin applied to rows};
\node[block, below=0.5cm of softmin] (logit) {logit applied entry-wise};
\node[block, below left=1.25cm of logit] (branch1) {Logits};
\node[block, below right=1.25cm of logit] (branch2) {Predictions};

\draw[arrow] (sequence) -- (embedding);
\draw[arrow] (embedding) -- (ssm);
\draw[arrow] (ssm) -- (distance);
\draw[arrow] (distance) -- (softmin);
\draw[arrow] (softmin) -- (logit);
\draw[arrow] (logit.south west) -- node[below, pos=0.5, xshift=30pt] {First output} (branch1.north);
\draw[arrow] (logit.south east) -- node[below, pos=0.5, xshift=-30pt] {Second output} (branch2.north);
\end{tikzpicture}
\end{center}

\section{Towards a mechanistic interpretation of COFFEE: auxiliary material}

\subsection{Simplified Version of the Induction Head Task: IH0}
\label{sec:reduction_IH}
To take a first step toward a mechanistic interpretability we simplify the induction head task. The solution of the problem by hand presents several challenges related to the embedding and state dimension, the sequence length, and the number of symbols in the vocabulary.

For this reason, we focus on addressing a toy version of the induction head task. We set:
\begin{enumerate}
    \item Embedding dimension $D=2$ and state dimension $n=1$
    \item Sequence length $L_{\text{seq}}=4$
    \item Vocabulary made of three symbols $V=\{1,2,3\}$
\end{enumerate}
An embedding in $\mathbb{R}^{2}$ allows us to employ only two one-dimensional state-space models. It is hence possible to visualize the trajectories of both states on a Cartesian plane. The choice of setting $L_{\text{seq}} = 4$ and using a vocabulary consisting of only three symbols, 
significantly reduced the number of possible sequences that could be generated according to the rules described in Appendix \ref{appendix:IH_task}. 

By specilizing COFFEE as in Equation~(\ref{eq:final_model}) for $D = 2$ and $n = 1$, we obtain the following setting:
\begin{equation}
\label{for:simplilfied_model}
\Sigma^{(i)} =
\left\{
\begin{array}{l}
  x(k)^{(i)} = (1+\lambda^{(i)}\Delta_i(k))
  x(k-1)^{(i)} + \Delta_i(k)
  u(k)_{i} \\
  y(k)^{(i)} = C^{(i)}x(k)^{(i)}
\end{array}
\right.
\end{equation}
with:
\begin{equation*}
\Delta_i(k) = \sigmoid(W_D^{(i)} x(k-1)^{(i)})
\end{equation*}
with $i=0,1$; $x(k)^{(i)} \in \mathbb{R}$ and $W_D^{(i)} \in \mathbb{R}$. The possible sequences are now of the form:
\[
\text{trigger}\ \|\ \ \text{target}\ \|\ \ \text{noise}\ \|\ \ \text{trigger}
\quad\textrm{or}\quad
\text{noise}\ \|\ \ \text{trigger}\ \|\ \ \text{target}\ \|\ \ \text{trigger}
\]
where, the trigger, target and noise subsequences are all of length one, resulting in a total sequence length of $L_{\text{seq}} = 4$. 
With these choices, the number of possible input sequences is reduced to just 8, allowing us to check the validity of a proposed model by inspection.

\subsection{COFFEE simplified model and state evolution for IH0}\label{sec:seqev}
We illustrate how the reduced COFFEE model, introduced to illustrate the functioning principles of the method, is able to solve IH0 by depicting the learned encoded symbols and the state evolution corresponding to a given input sequences. By setting $\lambda^{(i)} = 0$ and $C^{(i)} = 1$ for all $i = 0, 1$,  each \textbf{state} $x(k)^{(i)}$ becomes equal to the corresponding output $y(k)^{(i)}$, and the system behaves as a perfect \textbf{integrator}. This is desirable to our aim since it implments a perfect dynamical memory in absence of inputs. The model thus becomes:
\begin{equation}
\label{for:simplilfied_model_for_hand_sol}
\Sigma^{(i)} =
\left\{
\begin{array}{l}
  x(k)^{(i)} = x(k-1)^{(i)} + \Delta_i(k)
  u(k)_{i} \\
  y(k)^{(i)} = x(k)^{(i)}
\end{array}
\right.
\end{equation}
with:
\begin{equation*}
\Delta_i(k) = \sigmoid(W_D^{(i)} x(k-1)^{(i)})
\end{equation*}
where $x(k)^{(i)} \in \mathbb{R}$ and $W_D^{(i)} \in \mathbb{R}$ for $i = 0, 1$. 
We set $W_D^{(i)} = 1$ for $i=0,1$, in order to allow for more direct interpretability of the effect of state feedback in this setting.

The only parameters to be learned at this point are embedding vectors for the symbols in the vocabulary $V = \{1, 2, 3\}$. According to the principles {\bf P1-P3} described in Section \ref{sec:theoretical_insights}, we start with the embedding parameters initialization:
\begin{eqnarray*}
1 & \rightarrow & [6,6]^T\\
2 & \rightarrow & [-10,-1]^T\\
3 & \rightarrow & [-1,-10]^T.
\end{eqnarray*}
Recall that, without loss of generality, symbol ``1'' is assumed to be the trigger.
The final, learned solution is:
\begin{eqnarray*}
1 & \rightarrow & [5.394,5.343]^T\\
2 & \rightarrow & [-10.264,-1.575]^T\\
3 & \rightarrow & [-1.539,-10.340]^T.
\end{eqnarray*}
Such solution can provably solve correctly the problem with 100\% accuracy. This can be verified by inspection, since the possible input sequences are just 8.

In fact, with these parameter choices and the introduced simplifications, it is manageable to visualize how the possible input sequences are processed by the simplified model. All vectors are represented in $R^2,$ and we use $z_1,z_2$ to denote the abscissa and ordinate, respectively (we avoid using $x,y$ to avoid confusion with the state $x(k)$ and the output $y(k)$). Specifically, the outputs of $\Sigma^{(0)}$ generate the $z_1$-coordinates of the output vectors, while the outputs of $\Sigma^{(1)}$ generate the $z_2$-coordinates.

A picture illustrating how the input sequence ``$1 \| 2 \| 3 \| 1$'' is processed is provided next, which also illustrates the intuitive motivation for the particular choice of parameters. It is instructive to compare it with the trajectory associated to the sequence ``$3 \| 1 \| 2 \| 1$'', in order to see how the different positioning of the trigger with respect to the noise influences the dynamics.

\begin{figure}[H]
\begin{center}
\begin{tikzpicture}[scale=0.5,every node/.style={scale=0.8}, >=stealth, thick]
\draw[->] (-11.5, 0) -- (7.5, 0) node[right] {$z_1$};
\draw[->] (0, -11.5) -- (0, 6) node[above] {$z_2$};

\foreach \x in {-10,-5,5} {
    \draw (\x,0.15) -- (\x,-0.15) node[below,yshift=-2pt] {\scriptsize $\x$};
}

\foreach \y in {-10,5} {
    \draw (0.15,\y) -- (-0.15,\y) node[left,xshift=-2pt] {\scriptsize $\y$};
}

\fill[ForestGreen,opacity=0.075] (0,0) -- (-0.447*6.5,0.894*6.5) -- (-11.5,0.894*6.5) -- (-11.5,-11.5) -- cycle;
\fill[Orange,opacity=0.075] (0,0) -- (0.892*8.25,-0.452*8) --  (0.892*8.25,0.894*6.5) -- (-0.447*6.5,0.894*6.5) -- cycle;
\fill[Blue,opacity=0.075] (0,0) -- (0.892*8.25,-0.452*8) -- (0.892*8.25,-11.5) -- (-11.5,-11.5) -- cycle;

\node[ForestGreen] at (-10,5) {Region ``2''};
\node[Orange] at (1.75,5) {Region ``1''};
\node[Blue] at (1.75,-10.75) {Region ``3''};

\draw[->,Black!60] (0, 0) -- (5.39, 5.34) node[xshift=0.6cm, yshift=-0.6cm] {$``1"$, \scriptsize $\left( \begin{matrix}
    5.39 \\
    5.34
\end{matrix} \right)$};

\draw[->,Black!60] (0, 0) -- (-10.26, -1.57) node[below=0.15cm] {$``2"$, \scriptsize $\left( \begin{matrix}
    -10.26 \\
    -1.57
\end{matrix} \right)$};

\draw[->,Black!60] (0, 0) -- (-1.54, -10.34) node[left=0.15cm] {$``3"$, \scriptsize $\left( \begin{matrix}
    -1.54 \\
    -10.34
\end{matrix} \right)$};


\draw[->,color = Turquoise] (0, 0) -- (2.69, 2.67) node[midway,below right] {$k=0$};

\draw[->,color = Turquoise] (2.69, 2.67) -- (-6.92, 1.2) node[midway,above=0.15cm] {$k=1$};

\draw[->,color = Turquoise] (-6.92, 1.2) -- (-6.92,-6.74) node[midway,right=0.15cm] {$k=2$};

\draw[->,color = Turquoise] (-6.92,-6.74) -- (-6.92,-6.73) node[midway,left=0.15cm] {$k=3$, \scriptsize $\left( \begin{matrix}
  -6.92 \\
  -6.73
\end{matrix} \right)$}; 

\draw[->,color = BrickRed] (0, 0) -- (-0.77,-5.17) node[midway, left=0.15cm] {$k=0$};

\draw[->,color = BrickRed] (-0.77,-5.17) -- (0.92,-5.1) node[midway,above=0.15cm] {$k=1$};

\draw[->,color = BrickRed] (0.92,-5.1) -- (-6.42,-5.14) node[midway,above=0.15cm] {$k=2$};

\draw[->,color = BrickRed] (-6.42,-5.14) -- (-6.41,-5.11) node[midway,below right=0.15cm] {$k=3$, \scriptsize $\left( \begin{matrix}
  -6.41\\
  -5.11
\end{matrix} \right)$};
\end{tikzpicture}
\end{center}
\caption{Embeddings (in grey) and trajectories of the state for the reduced model solving the IH0 problem, with input $1\|2\|3\|1$ and $3\|1\|2\|1$ (in blue and red, respectively). The last trigger does not move significantly the state since the sigmoids in the input gates are not activated. The colored regions indicate the sections of the plane corresponding to each possible choice of the final output symbol (e.g final states in Region 2 lead to output ``2'').}
\label{fig:trajectory_of_hand_crafted_sol}
\end{figure}
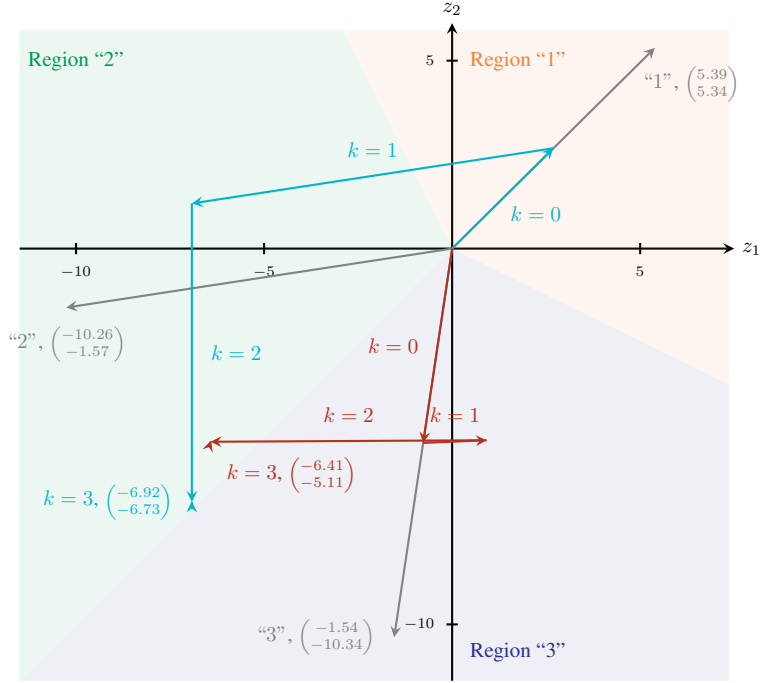
In the above diagram the \textbf{arrow heads} of the light blue vectors point to the output $y(k) \in \mathbb{R}^{2}$ (and also the state $x(k) \in \mathbb{R}^{2}$, because $y(k)^{(i)} = x(k)^{(i)}, i=0,1$) of the simplified SSM model at the time steps $k = 0, 1, 2, 3$), whose coordinates $\left( \begin{matrix}
    z_1 \\
    z_2
\end{matrix} \right) \in \mathbb{R}^{2}$ are also provided.
Vectors in gray are the embedding vectors of the symbols in $V$ with their coordinates.

The qualitative behavior for the ``$1 \| 2 \| 3 \| 1$'' sequence is as expected: the trigger moves the state in an activation area; the target symbol moves it towards its own representation, where the gate activation is also lower; the subsequent symbols do not move the state outside of the correct decision area. For the other sequence, after the first noise symbol moves the state in an area of low activation for the second state variable, the dynamics plays as expected but mostly along the first variable (horizontally). All other sequences have similar behavior to the ones depicted.

\section{Results on the Induction Head Task (IH)}
In the subsequent paragraphs, additional details on the training procedure for the induction head task are given. Moreover, initialization details are provided for both the COFFEE model and S6.
\label{Appendix:result_IH}

\subsection{Experiment settings}
We do not rely on a predefined dataset with a fixed set of sequences. For instance, by setting the sequence length to \( L_{\text{seq}} = 16 \) and defining the vocabulary as \( V = \{1, 2, 3, 4, 5, 6, 7\} \), the total number of possible sequences that can be generated, according to the rules described in Appendix \ref{appendix:IH_task}, exceeds several hundred billion.

To address this, we developed a software tool capable of generating such sequences randomly on demand. We created two independent instances of this tool to generate training and validation sequences.

Although it is theoretically possible for the same sequence to appear in both the training and validation sets, this is extremely unlikely given that the total number of training sequences used is capped at 512 million.

As the loss function, we employed \textbf{cross-entropy}, and for optimization, we used the \textbf{Adam} optimizer \cite{Adam}. The learning rate is specified for each experiment individually.

In each training iteration, we randomly generate and extract 512 sequences. This process is repeated 10'000 times, which we define as one \textbf{epoch}. The maximum number of epochs is set to 100. However, if good performance is achieved before reaching this limit, the training process is stopped early.

At the end of each epoch, we evaluate the model's generalization ability on a randomly generated set of 10'000 sequences, computing the \textbf{loss} and/or \textbf{accuracy}. This evaluation also allows us to save the model that performs best during training.

\subsection{Models Initialization}
In this section, we provide details about the initialization of our model and the S6 model used in our experiments.
\subsection*{Our model}
The parameters of our model are initialized as follows:
\begin{enumerate}
    \item Transition matrices \( A^{(i)} \in \mathbb{R}^{n \times n} \). These matrices are diagonal, with entries initialized to \(0\);
    \item State-to-output row vectors \( C^{(i)} \in \mathbb{R}^{n} \), whose entries are initialized according to the standard normal distribution \(\mathcal{N}(0, 1)\);
    \item The vectors \( W_D^{(i)} \in \mathbb{R}^{n} \), whose entries are also initialized according to the standard normal distribution \(\mathcal{N}(0, 1)\).
\end{enumerate}
Regarding the embedding matrix, we initialize it as follows:
\begin{enumerate}
    \item We generate a random matrix \( L \in \mathbb{R}^{D \times |M|} \) with entries sampled from a uniform distribution in the interval \([0, 1)\);
    \item The matrix \( L \) is then decomposed using QR decomposition as \( L = QR \), where \( Q \in \mathbb{R}^{D \times |M|} \) has all the columns \textbf{orthogonal} to each other and of \textbf{norm} one;
    \item Finally, we take \( Q^{T} \in \mathbb{R}^{|M| \times D} \) and set it as the initial embedding matrix. This procedure yields an orthonormal initialization of the embedding vectors.
\end{enumerate}

\subsection*{S6 model}
The parameters in S6 are initialized as follows:
\begin{enumerate}
    \item Transition matrices \( A^{(i)} \in \mathbb{R}^{n \times n} \). These matrices are diagonal, with entries initialized according to HiPPo theory \cite{HiPPo}. In particular, the \(i\)-th diagonal entry is initialized as \( -(i+1),i=0,...,n-1 \). For more details see \cite{MAMBA}, sec. 3.6 pag. 9;
    \item The weight matrix \( W_B \in \mathbb{R}^{n \times D} \) is initialized according to the standard normal distribution \(\mathcal{N}(0, 1)\);
    \item The weight matrix \( W_C \in \mathbb{R}^{n \times D} \) is initialized according to the standard normal distribution \(\mathcal{N}(0, 1)\);
    \item The weight matrix \( W_D \in \mathbb{R}^{D \times D} \) is initialized according to the standard normal distribution \(\mathcal{N}(0, 1)\);
    \item The embedding matrix in \(\mathbb{R}^{|M| \times D}\), is initialized according to the standard normal distribution \(\mathcal{N}(0, 1)\).
\end{enumerate}

\section{MNIST Architecture and Results}
\label{Appendix:MNIST_arch_and_results}
In this section, we describe the experimental setup used to obtain the results in Section~\ref{sec:MNIST}. Specifically, we (i) describe the MNIST dataset, (ii) a limitation of the sMNIST task in evaluating SSM performance, and (iii) present the employed architecture to address this classification task.
\subsection{MNIST Dataset}
The MNIST dataset \cite{lecun2010mnist} is a widely adopted benchmark in machine learning, particularly for image classification tasks. It comprises 70'000 grayscale images of handwritten digits from 0 to 9, each with a resolution of $28 \times 28$ pixels. The dataset is split into a training set of 60'000 images and a test set of 10'000 images. Thanks to its simplicity and standardized format, MNIST serves as a common starting point for evaluating new models and algorithms.

In our experiments, we construct a validation set by randomly sampling 10'000 images from the original \textsc{MNIST} training set. Consequently, the dataset is partitioned into 50'000 training images, 10'000 validation images, and 10'000 test images. Moreover, we did not use the full $28 \times 28$ pixel images. Instead, we opted to crop them to $25 \times 25$ pixels.

\subsection{Architecture Description}
In this section, we first discuss a limitation in assessing the true performance of an SSM layer in solving the \textbf{sMNIST} task, a variant commonly used to evaluate state-space models performance. We then present our architecture and its subsequent developments. Our approach allows for a more accurate evaluation of the effectiveness of state-space models.

\subsubsection{Limitations of the sMNIST Task}
\label{subsec:limit_sMNIST}
A review of the literature shows that, in general, to assess the ability of state space models to capture long-range dependencies, they are tested on a modified version of the MNIST task called \textbf{sMNIST}, which stands for \textbf{sequential MNIST}. In this task, each $28 \times 28$ pixel image is vectorized. Specifically, the image is converted into a column vector in $\mathbb{R}^{784}$ by vertically stacking all columns of size $\mathbb{R}^{28}$, a method known as column-major order. Alternatively, the same result can be achieved by horizontally stacking all rows of the image to form a vector in $\mathbb{R}^{784}$, known as row-major order.

Drawing inspiration from the literature, we initially attempted to solve the \textbf{sMNIST} task using the following architecture:
\begin{center}
\begin{tikzpicture}[
  node distance=1cm and 2cm, 
  block/.style={rectangle, draw, minimum width=3cm, minimum height=1cm, align=center},
  arrow/.style={-{Latex}, thick},
  scale=0.75,every node/.style={scale=0.75},
]

\node[block] (input_seq) {Image as a column vector in $\mathbb{R}^{784}$};
\node[block, below=0.75cm of input_seq] (ssm_layer) {Single SISO SSM layer};
\node[block, below=0.75cm of ssm_layer] (gelu) {GELU non linearity};
\node[block, below=0.75cm of gelu] (linear_layer) {Linear layer: $\mathbb{R}^{784} \rightarrow \mathbb{R}^{10}$};
\node[block, below=0.75cm of linear_layer] (prediction) {$\argmax$};

\draw[arrow] (input_seq) -- (ssm_layer);
\draw[arrow] (ssm_layer) -- (gelu);
\draw[arrow] (gelu) -- (linear_layer);
\draw[arrow] (linear_layer) -- (prediction);
\end{tikzpicture}
\end{center}
With this architecture, we take as input the vectorized image in $\mathbb{R}^{784}$. The image, now represented as a sequence of pixels, is processed by an SSM layer parameterized with our model. A GELU \cite{gelu} activation function is applied to each element of the resulting output sequence.\footnote{The Gaussian Error Linear Unit (GELU) activation function is defined as $ \mathrm{GELU}(x) = x \cdot \Phi(x)$, where \(\Phi(x)\) is the cumulative distribution function (CDF) of the standard normal distribution. For more details, see \cite{gelu}.} After the GELU, we still have a vector in $\mathbb{R}^{784}$, which is mapped to $\mathbb{R}^{10}$ by a linear layer. This linear layer produces the logits for each of the ten classes. The predicted class is the one with the highest logit, obtained by taking the $\argmax$ over the entries of the resulting vector in $\mathbb{R}^{10}$. Specifically,
\[
\hat{i} = \argmax_{i \in \{0,...,9\}} \quad v_i
\]
where with $\hat{i}$ we denote the predicted class and with $v_i$ we denote the $i$-th entry of the vector $v \in \mathbb{R}^{10}$.

Each symbol in the input sequence represents, by construction, a pixel value. Consequently, each symbol is embedded as a scalar, resulting in a model dimension of $D=1$. By training the above architecture with a single state space model (since $D=1$) with:
\begin{enumerate}
    \item State dimension $n=8$;
    \item $100$ epochs;
    \item Batch size $512$;
    \item Using an initial learning rate of $\eta=0.01$, which was reduced to $\eta=0.005$ once the training loss fell below $0.150$;
    \item Data augmentation, using a \textbf{roto-translation} with a maximum rotation of $5$ degrees and a maximum translation of $0.01$ along both the $x$ and $y$ axes \footnote{For instance, if we set translate=(a, b) (i.e. $x$, $y$  translations respectively), then horizontal shift is randomly sampled in the range $[-\text{img}_{width} \cdot a,\text{img}_{width} \cdot a]$ and vertical shift is randomly sampled in the range $[-\text{img}_{height} \cdot b,\text{img}_{height} \cdot b$]. },
\end{enumerate}
we obtained a \textbf{loss} of $0.226$ and a \textbf{accuracy} of $93.6\%$. The employed loss function is the cross-entropy loss. To evaluate the contributions of the SSM layer and the linear layer to the observed performance, we trained the same architecture without the SSM layer, as illustrated in the following diagram:
\begin{center}
\begin{tikzpicture}[
  node distance=1cm and 2cm, 
  block/.style={rectangle, draw, minimum width=3cm, minimum height=1cm, align=center},
  arrow/.style={-{Latex}, thick},
  scale=0.75,every node/.style={scale=0.75},
]

\node[block] (input_seq) {Image as a column vector in $\mathbb{R}^{784}$};
\node[block, below=0.75cm of input_seq] (gelu) {GELU non linearity};
\node[block, below=0.75cm of gelu] (linear_layer) {Linear layer: $\mathbb{R}^{784} \rightarrow \mathbb{R}^{10}$};
\node[block, below=0.75cm of linear_layer] (prediction) {$\argmax$};

\draw[arrow] (input_seq) -- (gelu);
\draw[arrow] (gelu) -- (linear_layer);
\draw[arrow] (linear_layer) -- (prediction);
\end{tikzpicture}
\end{center}
we trained the architecture, again, with:
\begin{enumerate}
    \item $100$ epochs;
    \item Batch size $512$;
    \item An initial learning rate of $\eta=0.01$, which was reduced to $\eta=0.005$ once the training loss fell below $0.150$;
    \item Data augmentation implemented with a \textbf{roto-translation} with a maximum rotation of $5$ degrees and maximum translation of $0.01$ on both $x$ and $y$ axes;
\end{enumerate}
as a result, we obtained a \textbf{loss} of $0.284$ and a \textbf{accuracy} of $92.0\%$.

These results suggest that in this configuration the contribution of the SSM layer to the performance is negligible compared to that of the linear layer.

In light of this experiment, it becomes evident that, in order to properly assess the SSM's ability to solve the task, the use of linear layers should be minimized. Their strong effectiveness can obscure the contributions of state space models.

\subsubsection{Our Architecture}
Our model is designed as a general tool for sequence processing. In general, when dealing with sequences, we take a sequence of symbols from some vocabulary, associate an embedding vector to each symbol, and then process the sequence of embedding vectors with the model. For instance, this is done for the induction head task presented in Appendix \ref{appendix:IH_task}.  

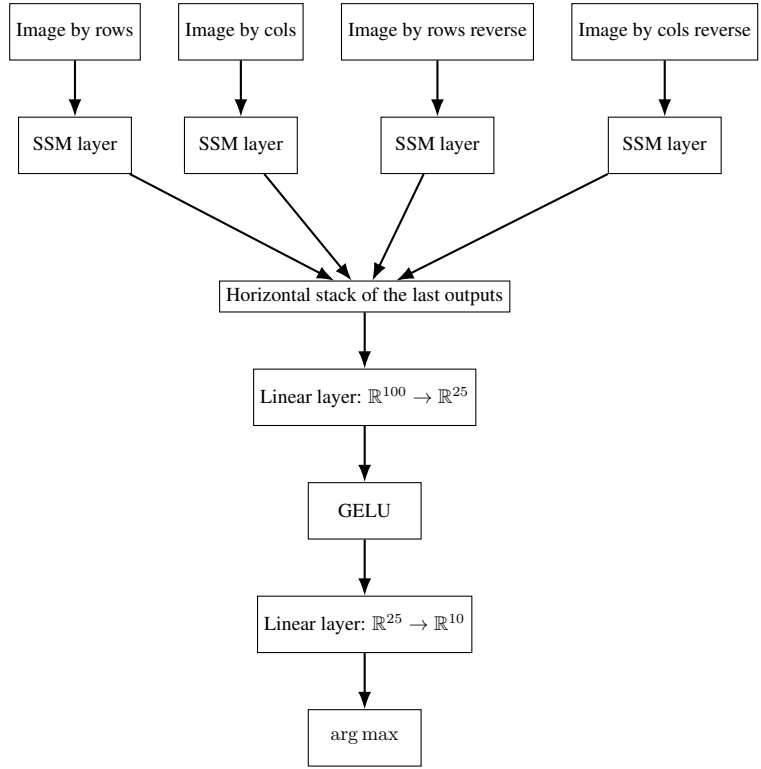
\begin{figure}[t]
\begin{center}
\begin{tikzpicture}[
  node distance=1cm and 0.5cm, 
  block/.style={rectangle, draw, minimum width=2cm, minimum height=1cm, align=center},
  arrow/.style={-{Latex}, thick},
  scale=0.75,every node/.style={scale=0.75}
]

\node[block] (input_seq1) {Image by rows};
\node[block, right=of input_seq1] (input_seq2) {Image by cols};
\node[block, right=of input_seq2] (input_seq3) {Image by rows reverse};
\node[block, right=of input_seq3] (input_seq4) {Image by cols reverse};
\node[block, below=0.75cm of input_seq1] (ssm_layer1) {SSM layer};
\node[block, below=0.75cm of input_seq2] (ssm_layer2) {SSM layer};
\node[block, below=0.75cm of input_seq3] (ssm_layer3) {SSM layer};
\node[block, below=0.75cm of input_seq4] (ssm_layer4) {SSM layer};
\node (horizontal_stack) [rectangle, draw, below=1.8cm of $(ssm_layer2)!0.63!(ssm_layer3)$] {Horizontal stack of the last outputs};
\node[block, below=0.75cm of horizontal_stack] (linear_layer1) {Linear layer: $\mathbb{R}^{100} \rightarrow \mathbb{R}^{25}$};
\node[block, below=0.75cm of linear_layer1] (gelu) {GELU};
\node[block, below=0.75cm of gelu] (linear_layer2) {Linear layer: $\mathbb{R}^{25} \rightarrow \mathbb{R}^{10}$};
\node[block, below=0.75cm of linear_layer2] (prediction) {$\argmax$};

\draw[arrow] (input_seq1) -- (ssm_layer1);
\draw[arrow] (input_seq2) -- (ssm_layer2);
\draw[arrow] (input_seq3) -- (ssm_layer3);
\draw[arrow] (input_seq4) -- (ssm_layer4);
\draw[arrow] (ssm_layer1) -- (horizontal_stack);
\draw[arrow] (ssm_layer2) -- (horizontal_stack);
\draw[arrow] (ssm_layer3) -- (horizontal_stack);
\draw[arrow] (ssm_layer4) -- (horizontal_stack);
\draw[arrow] (horizontal_stack) -- (linear_layer1);
\draw[arrow] (linear_layer1) -- (gelu);
\draw[arrow] (gelu) -- (linear_layer2);
\draw[arrow] (linear_layer2) -- (prediction);
\end{tikzpicture}
\caption{Proposed architecture.}
\label{fig:archs_MNIST}
\end{center}
\end{figure}

With the MNIST dataset, we want to adopt the same approach. As previously mentioned, we use cropped images of size $25 \times 25$ pixels. We treat an image as a sequence of \textbf{twenty-five} symbols (i.e. one for each row). For each symbol $s_i$, $i=0,\dots,24$, its embedding vector in $\mathbb{R}^{25}$ is given by the $i$-th row of the image. Thus, processing an image is equivalent to processing a sequence of twenty-five symbols embedded in $\mathbb{R}^{25}$. To clarify this idea, consider the following diagram.
\begin{center}
\begin{tikzpicture}[scale=0.75,every node/.style={scale=0.75}]
    \draw (-2,0) rectangle (6,3);

    \draw (-2,0.5) -- (6,0.5);
    \draw (-2,2.5) -- (6,2.5);

    \node at (2,2.75) {Embedding of the first symbol in $\mathbb{R}^{25}$};
    \node at (2,1.5) {$\vdots$};
    \node at (2,0.25) {Embedding of the last symbol in $\mathbb{R}^{25}$};

    \draw [->] (6.1,2.75) -- (7,2.75);
    \node[right] at (7,2.75) {First row of the image};
    \draw [->] (6.1,0.25) -- (7,0.25);
    \node[right] at (7,0.25) {Last row of the image};

    \node[left] at (-2,2.75) {$s_0$};
    \node[left] at (-2,0.25) {$s_{24}$};

    \draw[decorate, decoration={brace, amplitude=7.5pt, mirror}] 
        (-2,-0.1) -- (6,-0.1) 
        node[midway, yshift=-20pt]{Overall image};
    \draw[decorate, decoration={brace, amplitude=7.5pt, mirror}] 
        (-2.7,-0.1) -- (-2,-0.1) 
        node[midway, yshift=-20pt]{Symbols};
    
\end{tikzpicture}
\end{center}
Drawing inspiration from the celebrated two filter formula used for the smoothing problem, see \cite{smoothing_problem_Pavon} and \cite{smoothing_problem_Ferrante}, we can consider the image as a sequence of symbols $s_0 \| \dots \| s_{24}$, with each $s_i \in \mathbb{R}^{25}$, and build two sequences: one using the rows as embedding vectors and the other using the columns. Let $s_0 \| \dots \| s_{24}$ denote the sequence created from the rows and $s_0^{'} \| \dots \| s_{24}^{'}$ the one from the columns. The index $i$, $i=0,\dots,24$, can now be interpreted as \textbf{time}. By thinking of a sequence as a discrete time signal of finite support, given that we have access to all the signals in advance, we process the same image in four ways:
\begin{enumerate}
    \item By rows from $i=0$ to $i=24$;
    \item By columns from $i=0$ to $i=24$;
    \item By rows from $i=24$ to $i=0$;
    \item By columns from $i=24$ to $i=0$.
\end{enumerate}
The resulting architecture is illustrated in Fig.~\ref{fig:archs_MNIST}.

In this setup, we process the images by rows, columns, and by rows and columns in reverse order. Each SSM layer produces an output matrix in $\mathbb{R}^{25 \times 25}$, from which we consider only the last row. We then take the last rows of all four matrices and stack them horizontally to obtain a vector in $\mathbb{R}^{100}$. This vector is mapped to the corresponding logits using two linear layers separated by a GELU activation function. The predicted class is again selected as:
\[
\hat{i} = \argmax_{i \in \{0,...,9\}} \quad v_i
\]
where with $\hat{i}$ we denote the predicted class and with $v_i$ we denote the $i$-th entry of the vector $v \in \mathbb{R}^{10}$.
\newline
\newline
Let us now analyze the number of learnable parameters in this architecture:
\begin{enumerate}
    \item In Appendix \ref{appendix:derivation_of_our_model}, we see that an SSM layer has $3nD$ parameters, with $n$ being the state dimension and $D$ the embedding dimension. In our case, we use $n=2$ and $D=25$, so a single SSM layer has $150$ parameters. Since we have four of them, the total number of parameters for the SSM layers is $600$;
    \item The first linear layer, which maps vectors from $\mathbb{R}^{100}$ to $\mathbb{R}^{25}$, has $2500 + 25 = 2525$ parameters: $2500$ for the linear transformation and $25$ for the bias, resulting in an overall affine transformation;
    \item The second linear layer, which maps vectors from $\mathbb{R}^{25}$ to $\mathbb{R}^{10}$, has $250 + 10 = 260$ parameters: $250$ for the linear transformation and $10$ for the bias, resulting in an overall affine transformation.
\end{enumerate}
Taking all these contributions into account, the overall architecture has $3385$ parameters.
\newline
\newline
We trained this model with:
\begin{enumerate}
    \item State dimension $n=2$;
    \item $100$ epochs;
    \item Batch size $512$;
    \item An initial learning rate $\eta=0.01$ that was reduced to $\eta=0.005$ when the training loss fell below $0.450$;
    \item Data augmentation implemented with a roto-translation with a maximum angle of $5$ degrees and maximum translation of $0.01$ on both $x$ and $y$ axes.
\end{enumerate}
With these parameters, we obtain a \textbf{ loss} of $0.107$ and a \textbf{accuracy} of $96.6\%$ with \textbf{standard deviation} of 0.1 (repeated ten times).
\newline
\newline
For comparison with the S6 model, we trained the same architecture using the S6 model within the SSM layers instead of our model. As mentioned in Appendix \ref{appendix:derivation_of_our_model}, our model has fewer parameters than S6, so the total number of parameters in this case is $5885$. Initially, we tried with the same settings, namely:
\begin{enumerate}
    \item State dimension $n=2$;
    \item $100$ epochs;
    \item Batch size $512$;
    \item With an initial learning rate $\eta=0.01$ that was reduced to $\eta=0.005$ when the training loss fell below $0.450$;
    \item Data augmentation implemented with a roto-translation with a maximum angle of $5$ degrees and maximum translation of $0.01$ on both $x$ and $y$ axes,
\end{enumerate}
and we obtained a \textbf{loss} of $1.891$ with a \textbf{accuracy} of $28.2\%$ with \textbf{standard deviation} of 0.1 (repeated ten times).
\newline
\newline
We conducted a second experiment by increasing the state dimension for the S6 model. We trained the architecture with:
\begin{enumerate}
    \item State dimension $n=16$;
    \item $100$ epochs;
    \item Batch size $512$;
    \item With an initial learning rate $\eta=0.01$ that was reduced to $\eta=0.005$ when the training loss fell below $0.450$;
    \item Data augmentation implemented with a roto-translation with a maximum angle of $5$ degrees and maximum translation of $0.01$ on both $x$ and $y$ axes.
\end{enumerate}
In this case, the architecture has $10'085$ parameters, and we obtained a \textbf{loss} of $1.876$ with a \textbf{accuracy} of $28.6\%$ with \textbf{standard deviation} of 0.1 (repeated ten times).
\newline
\newline
Based on these results, we can conclude that our model achieves far superior performance compared to the S6 model, and this performance is obtained with fewer parameters. Moreover, our model is also competitive with the highly optimized models presented in \cite{prodsumnetMLP}, even without specific optimization.

\end{document}